\documentclass{article}

\usepackage{microtype}
\usepackage{graphicx}
\usepackage{subcaption}
\usepackage{booktabs} 
\usepackage{array}

\makeatletter
\DeclareRobustCommand\onedot{\futurelet\@let@token\@onedot}
\def\@onedot{\ifx\@let@token.\else.\null\fi\spacefactor1000}
\makeatother
\def\eg{\emph{e.g}\onedot} 
\def\ie{\emph{i.e}\onedot} 

\usepackage{pifont}
\usepackage{makecell} 
\usepackage[dvipsnames]{xcolor}

\usepackage{algorithm}
\usepackage{algpseudocode}
\newcommand{\KVSet}{\mathbf{KV}}

\usepackage{wrapfig}

\definecolor{cvprblue}{rgb}{0.21,0.49,0.74}
\definecolor{memory}{rgb}{0.82, 0.51, 0.50} 
\definecolor{current}{rgb}{0.49, 0.60, 0.74} 
\usepackage[pagebackref,breaklinks,colorlinks,allcolors=cvprblue]{hyperref}

\usepackage{hyperref}
\usepackage{xurl}

\newcommand{\tf}[1]{\textcolor{black}{#1}}
\newcommand{\wenq}[1]{\textcolor{black}{#1}}

\usepackage[accepted]{icml2026}

\usepackage{amsmath}
\usepackage{amssymb}
\usepackage{mathtools}
\usepackage{amsthm}

\usepackage[capitalize,noabbrev]{cleveref}

\theoremstyle{plain}

\theoremstyle{definition}

\theoremstyle{remark}

\usepackage[textsize=tiny]{todonotes}

\icmltitlerunning{WorldPlay: Towards Long-Term   Geometric Consistency for Real-Time Interactive World Modeling}

\begin{document}

\twocolumn[{
\renewcommand\twocolumn[1][]{#1}%

  \icmltitle{WorldPlay: Towards Long-Term   Geometric Consistency for Real-Time Interactive World Modeling}



  \icmlsetsymbol{equal}{*}
  \icmlaffilorder{hkust,beihang,tencent}
  \begin{icmlauthorlist}
    \icmlauthor{Wenqiang Sun}{equal,hkust,tencent}
    \icmlauthor{Haiyu Zhang}{equal,beihang,tencent}
    \icmlauthor{Haoyuan Wang}{equal,tencent}
    \icmlauthor{Junta Wu}{tencent}
    \icmlauthor{Zehan Wang}{tencent}
    \icmlauthor{Zhenwei Wang}{tencent}
    \icmlauthor{Yunhong Wang}{beihang}
    \icmlauthor{Jun Zhang}{hkust}
    \icmlauthor{Tengfei Wang}{tencent}
    \icmlauthor{Chunchao Guo}{tencent}
  \end{icmlauthorlist}

      \centering
 \vspace{2mm}
 \includegraphics[width=0.97\textwidth]{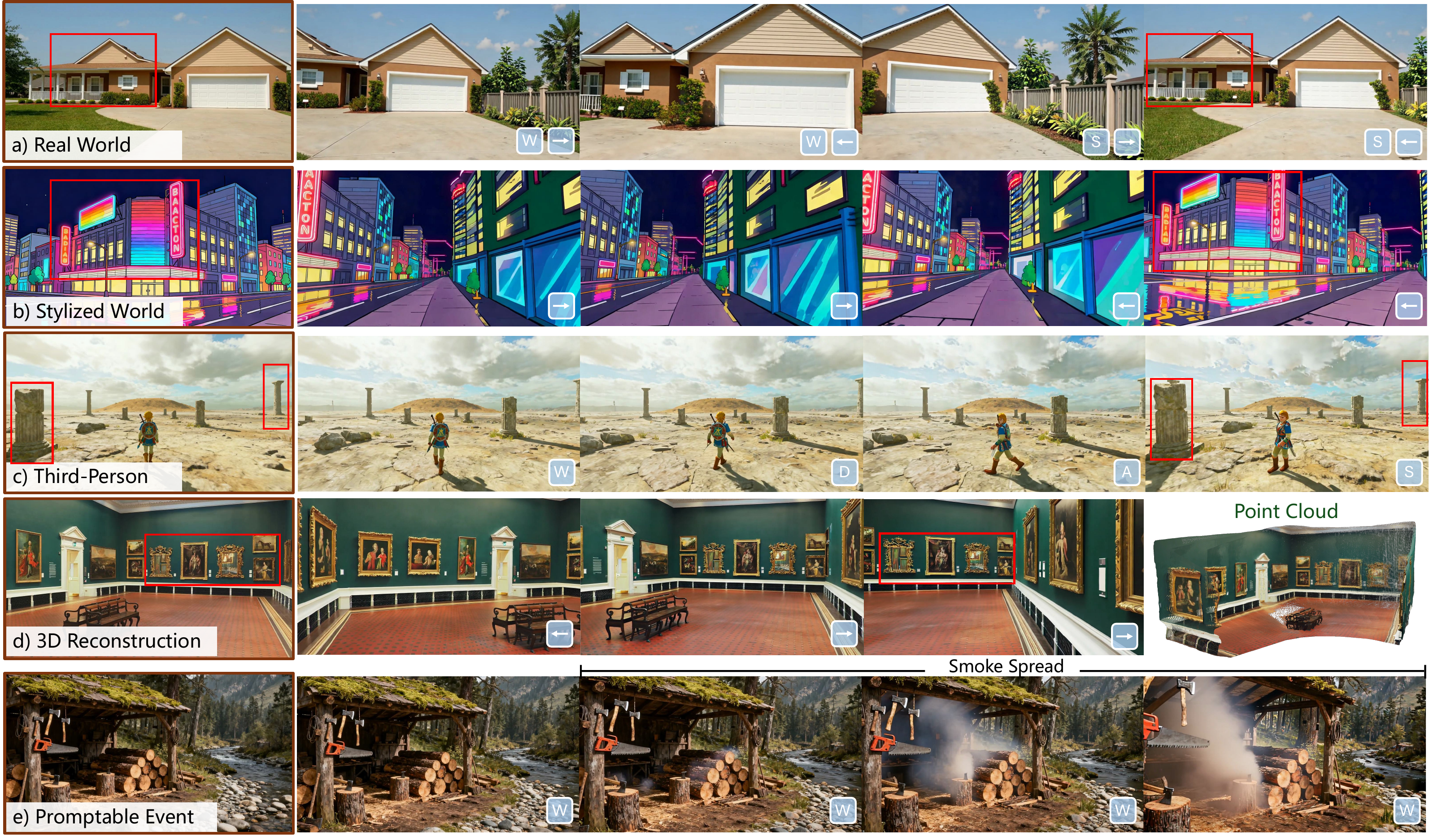}
    \vspace{-1mm}
  \captionof{figure}{\textbf{WorldPlay is a real-time, interactive world model that achieves long-term geometric consistency.} It responds to user navigation commands in a streaming fashion, while maintaining scenes remain coherent when revisiting \textit{(shown in red boxes)}. Our model shows remarkable generalization across diverse scenes, including \textbf{(a)} real world, \textbf{(b)} stylized world, and \textbf{(c)} third-person agent control. Furthermore, it supports \textbf{(d)} 3D scene generation via reconstruction and \textbf{(e)} dynamic world events triggered by text-based manipulation.}
    \label{fig:teaser}

  \icmlaffiliation{hkust}{Hong Kong University of Science and Technology}
  \icmlaffiliation{beihang}{Beihang University}
  \icmlaffiliation{tencent}{Tencent Hunyuan}

  \icmlcorrespondingauthor{Jun Zhang}{eejzhang@ust.hk}
  \icmlcorrespondingauthor{Tengfei Wang}{tengfeiwang12@gmail.com}
  \icmlcorrespondingauthor{Chuncaho Guo}{chunchaoguo@gmail.com}

  \icmlkeywords{Machine Learning, ICML}

  \vskip 0.3in
}]



\printAffiliationsAndNotice{\icmlEqualContribution}  

\begin{abstract}
\tf{This paper presents WorldPlay, a streaming video diffusion model that enables real-time, interactive world modeling with long-term geometric consistency, resolving the trade-off between speed and memory that limits current methods.  WorldPlay draws power from three key \wenq{ingredients}. 1) We use a Dual Action Representation to enable robust action control in response to the user's keyboard and mouse inputs. 2) To enforce long-term consistency, our Reconstituted Context Memory dynamically rebuilds context from past frames and uses  temporal reframing to keep geometrically important but long-past frames accessible, effectively alleviating memory attenuation.  3) We also propose Context Forcing, a novel distillation method designed for memory-aware model. Aligning memory context between the teacher and student preserves the student's capacity to use long-range information, enabling real-time speeds while preventing error drift. Taken together,  WorldPlay generates long-horizon streaming 720p video at 24 FPS with superior consistency, comparing favorably with existing techniques and showing strong generalization across diverse scenes.
\wenq{Project page and online demo can be found: \href{https://3d-models.hunyuan.tencent.com/world/}{https://3d-models.hunyuan.tencent.com/world/} and \href{https://3d.hunyuan.tencent.com/sceneTo3D}{https://3d.hunyuan.tencent.com/sceneTo3D}.}}
\end{abstract}    
\section{Introduction}
\label{sec:intro}
\tf{World models are driving a pivotal shift in computational intelligence, moving beyond language-centric tasks towards visual and spatial reasoning. By simulating dynamic 3D environments, these models empower agents to perceive and interact with complex surroundings, opening up new possibilities for embodied robotics and game development.}

\tf{At the forefront of world modeling is real-time interactive video generation, which aims at autoregressively predicting future video frames (or \textit{chunks}) to deliver instant visual feedback in response to every user's keyboard command. Despite significant progress, a fundamental challenge persists: \textit{how to simultaneously achieve real-time generation (speed) and long-term geometric consistency (memory)} in interactive world modeling. One class of methods~\cite{oasis,parkerholder2024genie2,he2025matrix}  prioritizes speed with distillation but neglects memory, resulting in inconsistency where scenes change upon revisit. The other class preserves consistency with explicit~\cite{li2025vmem,ren2025gen3c} or implicit~\cite{xiao2025worldmem,yu2025context,chen2025learning} memory, but complex memory makes distillation non-trivial (Sec. \ref{sec:distill}). As summarized in Table~\ref{tab:compare_related_works},  the simultaneous achievement of both low latency and high consistency remains an open problem.}

\tf{To tackle this challenge, we develop \textbf{WorldPlay}, a \textit{real-time  and  long-term consistent  world model for  general scenes}. We consider this problem as a next chunk (16 frames) prediction task for generating streaming videos conditioned on actions from users. Building upon autoregressive diffusion models, WorldPlay draws power from the model’s three key ingredients below.
}

\tf{The first is \textbf{Dual Action Representation} for control over agent and camera movement. Previous works~\cite{oasis,parkerholder2024genie2,he2025matrix} typically rely on discrete keyboard inputs (\eg, W, A, S, D) as action signals, which afford plausible, scale-adaptive movement but suffer from ambiguity for memory retrieval that requires revisiting exact locations. Conversely, continuous camera poses ($R, T$) provide spatial locations but cause training instability due to scene scale variance in training data. To combine the best of both worlds, we  convert action signals into continuous camera poses and discrete keys, achieving robust control and accurate location caching.}

\tf{The second key design is \textbf{Reconstituted Context Memory} for maintaining long-term geometric consistency. We actively reconstitute the memory through a two-stage process,  moving beyond simple retrieval~\cite{yu2025context,xiao2025worldmem}. It first dynamically rebuilds a context set by querying past frames based on spatial and temporal proximity. To overcome the long-range decay (the fading influence of distant tokens in Transformers~\cite{su2024roformer}), we propose \textit{temporal reframing} to rewrite positional embeddings of these retrieved frames. This operation effectively ``pulls"  geometrically important but long-past memories closer in time, forcing the model to treat them as  recent. This process keeps the influence of relevant long-range information preserved, enabling robust free extrapolation with strong geometric consistency.}

\tf{The final key ingredient is \textbf{Context Forcing}, a novel distillation method designed for memory-aware models to enable real-time generation. Existing distillation methods~\cite{chen2024diffusion,huang2025self,yin2024improved} fail to keep long-term memory as there is a fundamental distribution mismatch: training a memory-aware autoregressive student to mimic a memory-less bidirectional teacher. Even when augmenting teacher with memory, mismatched memory context will cause distribution diverge. We solve this by aligning the memory context for teacher and student during distillation. This alignment facilitates effective distribution matching, enabling real-time speed without eroding the memory while alleviating error accumulation over long sequences.}

\tf{Taken together, WorldPlay achieves real-time, interactive video generation at 24 FPS (720p) while maintaining long-term geometric consistency under streaming user control. The model is built on a large-scale, curated dataset of 320K real and synthetic videos with a custom rendering and processing platform. As shown in Fig.~\ref{fig:teaser}, WorldPlay shows superior generation quality and remarkable generalization across diverse scenes including first- and third-person real and stylized worlds, and supports applications ranging from 3D reconstruction and promptable events.
}

\begin{table*}[!ht]
\caption{Comparison with recent interactive world models. WorldPlay distinguishes itself as a general-domain model that simultaneously achieves long-horizon video generation, flexible action control, real-time interactivity, and long-term geometric consistency. 'Con.' and 'Dis.' represent continue and discrete action, respectively.
    }
    \label{tab:compare_related_works}
    \vspace{-2mm}
\scriptsize
    \centering
    \renewcommand{\arraystretch}{1.5} 
    \newcommand{\gou}{\textcolor{ForestGreen}{\ding{52}}}
    \newcommand{\cha}{\textcolor{Red}{\ding{55}}}
    \begin{tabular}{>{\centering\arraybackslash}m{1.7cm}|
    >{\centering\arraybackslash}m{1.3cm}|
    >{\centering\arraybackslash}m{2.1cm}|
    >{\centering\arraybackslash}m{1.7cm}|
    >{\centering\arraybackslash}m{1.7cm}|
    >{\centering\arraybackslash}m{1.7cm}|
    >{\centering\arraybackslash}m{1.7cm}|
    >{\centering\arraybackslash}m{1.7cm}
}
        \toprule
        & \textbf{Yume}~\cite{mao2025yume} &\textbf{Matrix-Game 2.0}~\cite{he2025matrix} & \textbf{GameGenX}~\cite{che2024gamegen} & \textbf{GameCraft}~\cite{li2025hunyuan} & \textbf{WorldMem}~\cite{xiao2025worldmem} & \textbf{VMem}~\cite{li2025vmem}  & \textbf{WorldPlay} \\
        \hline
        Resolution & $544$p & $360$p & $720$p & $720$p & $360$p & $576$p  & $720$p\\
        \hline
        Action Space & Text & Dis. & Dis. & Con. & Dis. & Con.  & Con. + Dis. \\
        \hline
        Real-time & \gou & \gou & \cha & \cha & \cha & \cha  & \gou \\
        \hline
        Long-term Consistency & \cha & \cha & \cha & \cha & \gou & \gou  & \gou \\
        \hline
        Long-Horizon & \cha & \gou & \gou & \gou & \cha & \cha & \gou \\
        \hline
        Domain & General & General & General & General & Minecraft & Static Scene  & General \\
        \bottomrule
    \end{tabular}
    \vspace{-0.0cm}
    \vspace{-5mm}
\end{table*}

\section{Related Work}
\label{sec:related_work}

\textbf{Video Generation.} Diffusion models~\cite{ho2020denoising, lipman2022flow, song2020score} have emerged as the state-of-the-art approach in video generative modeling.
\cite{chen2024videocrafter2, guo2023animatediff, yang2024cogvideox} adopt the latent diffusion model (LDM)~\cite{rombach2022high} to learn video distribution in the latent space, achieving efficient video generation. 
Recently, autoregressive video generation models~\cite{chen2024diffusion, henschel2025streamingt2v, kim2024fifo} theoretically enable one to generate unlimited length videos, laying the foundation for world models.
With the advancement of powerful architectures~\cite{peebles2023scalable} and sophisticated data pipelines, ~\cite{veo, wan2025wan, kling, hailuo, gao2025seedance, kong2024hunyuanvideo}, which are trained on web-scale datasets, have demonstrated emergent zero-shot capabilities to perceive, model, and manipulate the visual world~\cite{wiedemer2025video}, making it feasible to simulate the physical world.

\noindent\textbf{Interactive and Consistent World Models.} World models aim to predict future states based on current and past states. 
Studies such as \cite{alonso2024diffusion, bar2025navigation, valevski2024diffusion, yu2025gamefactory, sun2025virtual, he2024cameractrl, wang2024motionctrl, miyato2023gta, kong2024eschernet, li2025cameras, bahmani2025ac3d, sun2024dimensionx, mao2025yume,mao2025yume2,xiang2025pan,tang2025hunyuan} adopt discrete, continuous action signals or text instructions to enable agents to navigate and interact with virtual environments. \cite{yesiltepe2025infinity} proposes a training-free framework for instruction-controllable video generation.
Subsequent works that aim to achieve geometric consistency can be categorized into two types: explicit 3D reconstruction and implicit conditioning. \cite{li2025vmem, yu2025wonderworld, ren2025gen3c, cao2025uni3c, yu2024viewcrafter, yu2025trajectorycrafter,liu2024reconx} ensure spatial consistency by explicitly reconstructing 3D representations and rendering condition frames from these representations. However, they heavily rely on reconstruction quality, making it challenging to maintain long-term consistency. Recent work~\cite{hunyuanworld2025tencent} constructs 3D world models explicitly, without relying on video generation models. Although achieving promising 3D generation results, they can not be performed in real-time. In contrast, \cite{xiao2025worldmem, yu2025context} achieve implicit conditioning by leveraging field-of-view (FOV) to retrieve relevant context from historical frames. Concurrent work~\cite{hong2025relic} achieves interactive generation with fixed-length consistency through context compression~\cite{zhang2025packing}. However, developing a real-time world model with long-horizon geometric consistency remains unsolved.

\noindent\textbf{Distillation.} 
For video diffusion models, existing approaches typically employ distillation \cite{salimans2022progressive, geng2025mean, frans2024one,li2025flash, zhang2025accvideo} to achieve few-step inference, achieving faster generation. 
\cite{sauer2024fast, sauer2024adversarial, kang2024distilling, lin2025diffusion, lin2024sdxl, lin2025autoregressive} adopt adversarial training strategies to enable few-step inference, however, they often suffer from training instability and mode collapse. 
\cite{yin2024one, yin2024improved, lu2025adversarial, shin2025motionstream} utilize Variational Score Distillation~\cite{wang2023prolificdreamer} to achieve outstanding few-step generation performance \wenq{in various tasks}. 
In addition, CausVid~\cite{yin2025slow} proposes distilling a causal student model from a bidirectional teacher diffusion model to achieve real-time autoregressive generation.
Furthermore, Self-Forcing~\cite{huang2025self} mitigates exposure bias by refining the rollout strategy of CausVid. Our method proposes context forcing to preserve both the interactivity and geometric consistency while achieving real-time generation.
\section{Method}

\begin{figure*}
  \centering
  \includegraphics[width=0.95\linewidth]{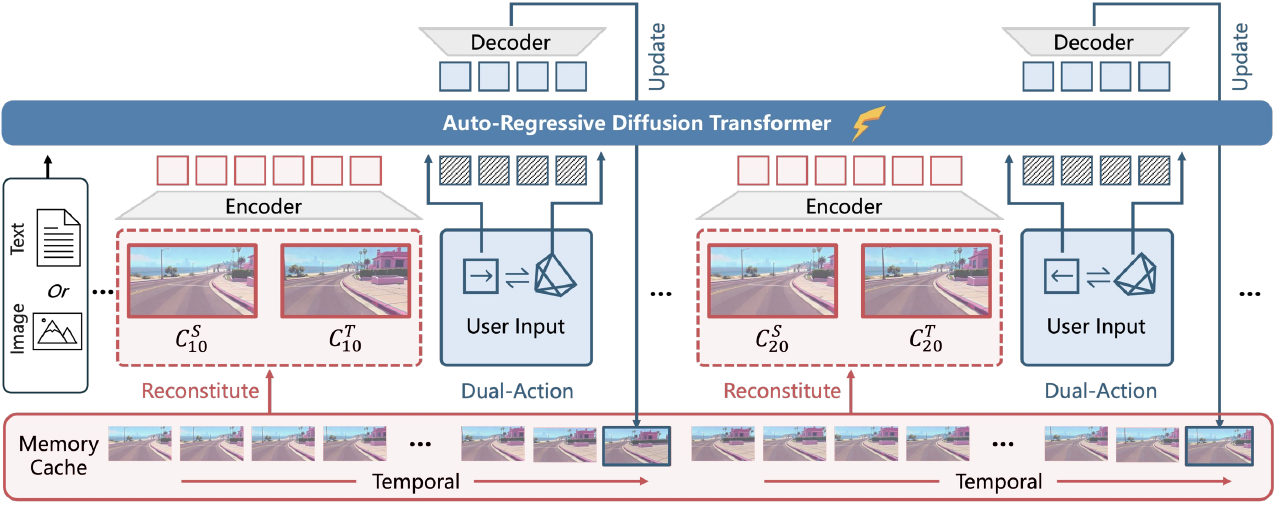}
  \vspace{-2mm}
  \caption{{\textbf{Method overview.} Given a single image or text prompt to describe a world, \textbf{WorldPlay} performs a next chunk (16 video frames)   prediction task to generate future videos conditioned on action from users. For the generation of each chunk, we dynamically reconstitute context memory from past chunks to enforce long-term temporal and geometric consistency.} 
  }
  \vspace{-3mm}
  \label{fig:pipeline}
\end{figure*}

Our goal is to construct a geometry-consistent and real-time interactive world model $N_{\theta}(x_{t}|O_{t-1}, A_{t-1}, a_{t}, c)$ parameterized by $\theta$, which can generate next chunk $x_{t}$ (a chunk is a few frames) based on past observations $O_{t-1}=\{x_{t-1}, ...,x_{0}\}$, action sequences $A_{t-1}=\{a_{t-1},...,a_{0}\}$, and current action $a_{t}$. Here, $c$ is a text prompt or image that describes the  world. For simplicity of notation, we omit $A, a, c$ in following sections. We first introduce the relevant preliminaries in Sec.~\ref{sec:pre}. In Sec.~\ref{sec:action}, we discuss the action representation for control. Sec.~\ref{sec:memory} describes our reconstituted context memory to ensure long-term geometric consistency, followed by Sec.~\ref{sec:distill} covering our context forcing, which mitigates exposure bias and enables few-step generation while maintaining long-term consistency. Finally, Sec.~\ref{sec:stream} details additional optimizations for real-time streaming generation. The pipeline is shown in Fig.~\ref{fig:pipeline}.

\subsection{Preliminaries}
\label{sec:pre}

\noindent\textbf{Full-sequence Video Diffusion Model.} Current video diffusion models~\cite{kong2024hunyuanvideo,wan2025wan} typically consist of a causal 3D VAE~\cite{kingma2013auto} and a Diffusion Transformer (DiT)~\cite{peebles2023scalable}, where each DiT block is composed of 3D self-attention, cross-attention, and feedforward network (FFN). 
The diffusion timestep is processed by positional embedding (PE) and a Multi-Layer Perceptron (MLP) to modulate the DiT blocks. The model is trained using flow matching~\cite{lipman2022flow}. Specifically, given a video latent $z_{0}$ encoded by the 3D VAE, a random noise $z_{1}\sim \mathcal{N}(0,I)$, and a diffusion timestep $k\in[0,1]$, an intermediate latent $z_{k}$ is obtained through linear interpolation. The model is trained to predict the velocity $v_{k}=z_{0}-z_{1}$,
\begin{equation}
     \mathcal{L}_{\text{FM}}(\theta) 
     =\mathbb{E}_{k, z_{0}, z_{1}} \Big\|N_{\theta}(z_k,k) - v_{k} \Big\|^2. 
     \label{training}
\end{equation}

\begin{figure}
  \centering
  \includegraphics[width=0.97\linewidth]{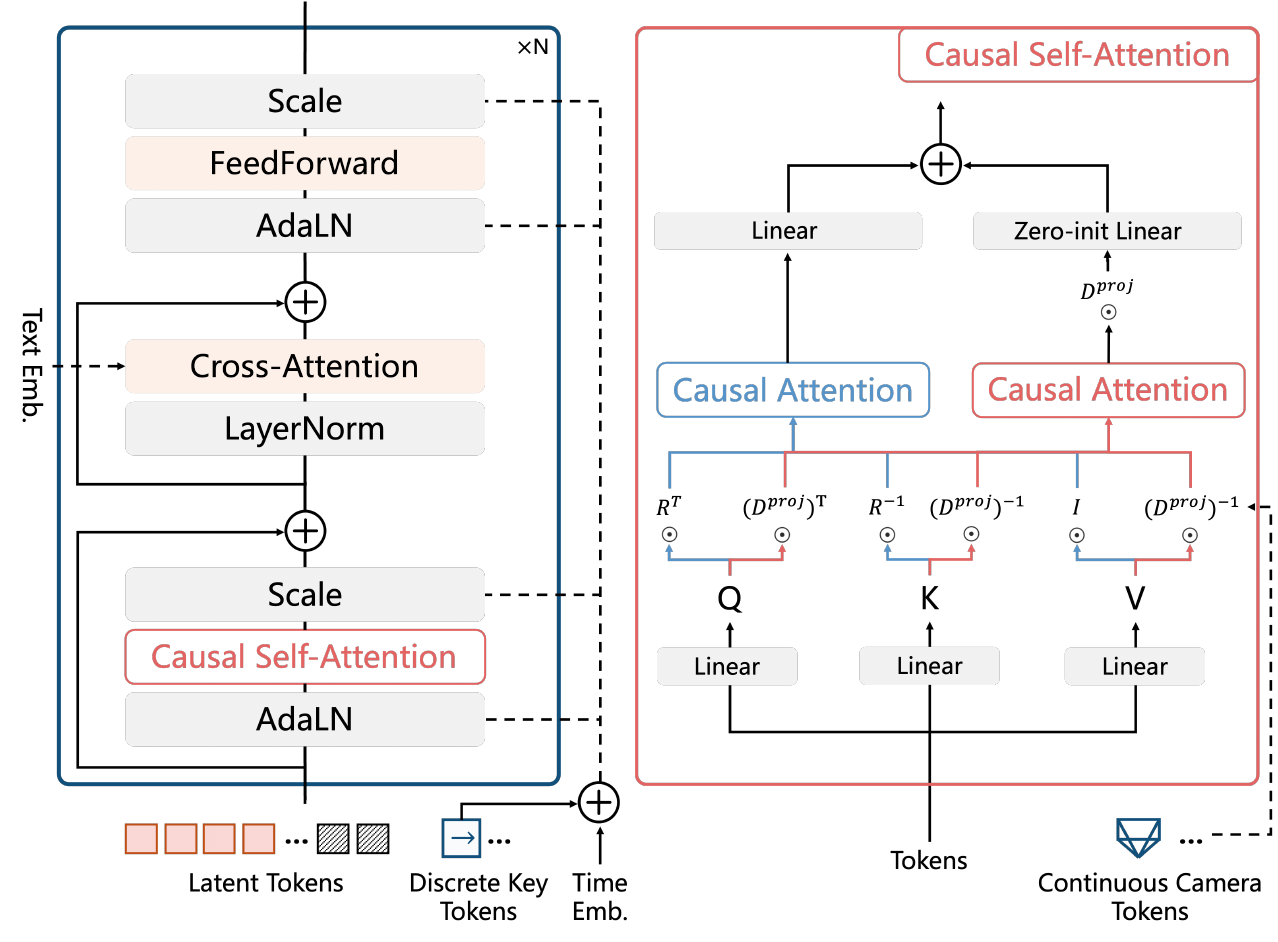}
  \vspace{-2mm}

  \caption{{Detailed architecture of our autoregressive diffusion transformer. The discrete key is incorporated with time embedding, while the continuous camera pose is injected into causal self-attention through PRoPE~\cite{li2025cameras}.} 
  }
  \label{fig:architecture}
  \vspace{-4mm}
\end{figure}

\noindent\textbf{Chunk-wise Autoregressive Generation.} However, the full-sequence video diffusion model is a non-causal architecture, which limits its ability for infinite-length interactive generation. Inspired by Diffusion Forcing~\cite{chen2024diffusion}, we finetune it into a chunk-wise autoregressive video generation model. Specifically, for video latent $z_{0} \in \mathbb{R}^{C\times T\times H\times W}$, we divide it into $\frac{T}{4}$ chunks $\{z_{0}^{i}\in \mathbb{R}^{C\times 4\times H\times W}|i=0,...,\frac{T}{4}-1\}$, and thus each chunk (4 latents) can be decoded into 16 frames. During training, we add different noise levels $k_{i}$ for each chunk and modify the full-sequence self-attention to block causal attention. The training loss is similar to Eq.~\ref{training}.

\subsection{Dual Action Representation for Control}
\label{sec:action}

Existing methods use keyboard and mouse inputs as action signals and inject the action control via MLP~\cite{oasis, xiao2025worldmem} or attention blocks~\cite{he2025matrix, yu2025context}. This enables the model to learn physically plausible movements across scenes with diverse scales (\eg very large and small scenes). However, they struggle to provide precise previous locations for spatial memory retrieval. In contrast, camera poses (rotation matrix and translation vector) provide accurate spatial locations that facilitate precise control and memory retrieval, but training only with camera poses faces challenges in training stability due to the scale variance in the training data. To address this, we propose a dual action representation that combines the best of both worlds as shown in Fig.~\ref{fig:architecture}. This design not only caches spatial locations for our memory module in Sec.~\ref{sec:memory}, but also enables robust and precise control. Specifically, we employ PE and a zero-initialized MLP to encode discrete keys and incorporate it into the timestep embedding, which is then used to modulate the DiT blocks. For continuous camera pose, we leverage relative positional encoding, \ie, PRoPE~\cite{li2025cameras}, which offers greater generalizability than commonly used raymaps, to inject complete camera frustums into self-attention blocks. The original self-attention computation is as follows,
\begin{equation}
     Attn_{1}=Attn(R^\top \odot Q, R^{-1} \odot K, V),
\end{equation}
\noindent where $R$ represents the 3D rotary PE (RoPE)~\cite{su2024roformer} for video latents. To encode frustum relationships between cameras, we utilize an additional attention,
\begin{equation}
    \begin{split}
     Attn_{2}= & D^{proj} \odot Attn((D^{proj})^\top \odot Q, \\
     &(D^{proj})^{-1} \odot K, (D^{proj})^{-1} \odot V),
    \end{split}
\end{equation}
\noindent here, \wenq{$D^{proj}=\begin{bmatrix}
        K & \mathbf{0}\\
        \mathbf{0} & 1
    \end{bmatrix}
    T^{cw}$ is derived from the camera's intrinsic $K$ and extrinsic parameters $T^{cw}$}, as described in \cite{li2025cameras}. The result of each self-attention block is $Attn_{1} + zero\_init(Attn_{2})$.

\subsection{Reconstituted Context Memory for Consistency}
\label{sec:memory}

\begin{figure}
  \centering
  \begin{subfigure}{0.32\linewidth}
    \includegraphics[width=0.95\linewidth]{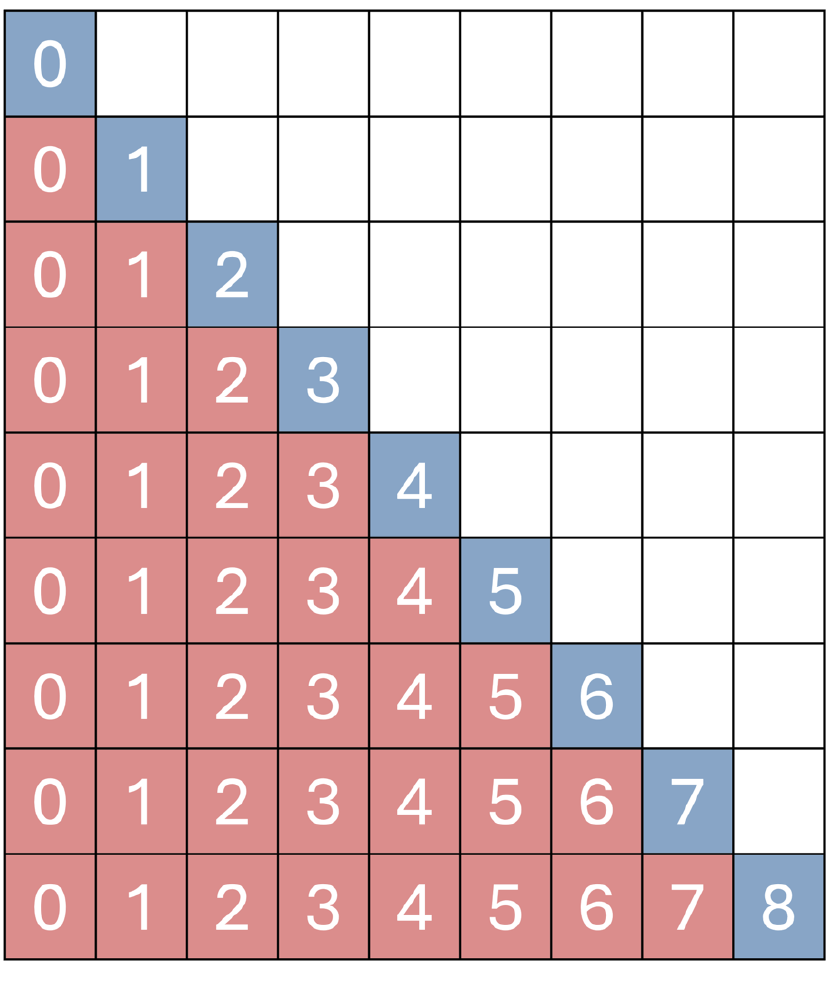}
    \caption{Full context}
    \label{fig:attn-a}
  \end{subfigure}
  \begin{subfigure}{0.32\linewidth}
    \includegraphics[width=0.95\linewidth]{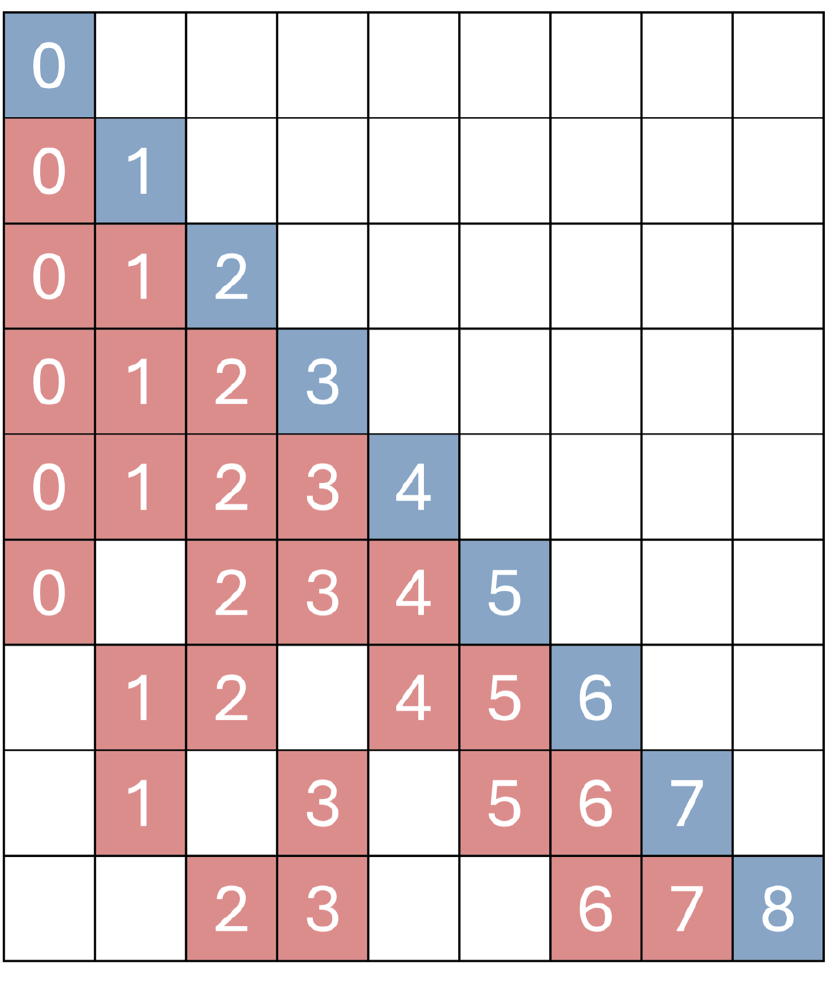}
    \caption{Absolute indices}
    \label{fig:attn-b}
  \end{subfigure}
  \begin{subfigure}{0.32\linewidth}
    \includegraphics[width=0.95\linewidth]{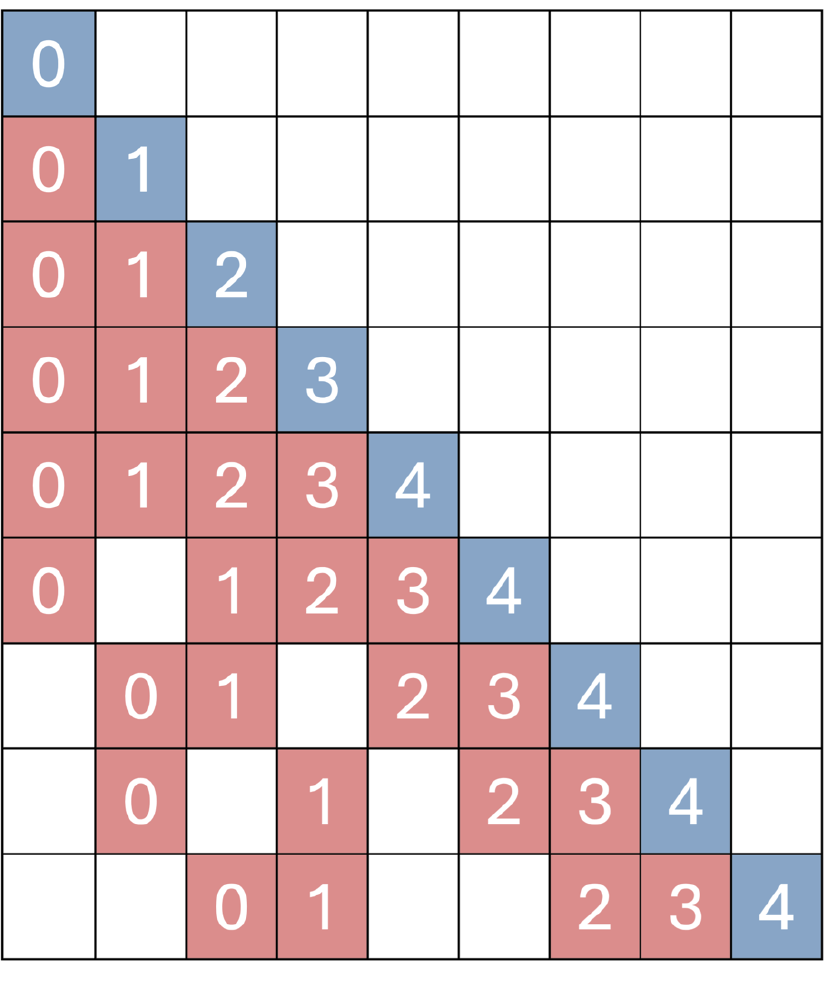}
    \caption{Relative indices}
    \label{fig:attn-c}
  \end{subfigure}
  \vspace{-2mm}
  \caption{\textbf{Memory mechanism comparisons}. The \textcolor{memory}{red} and \textcolor{current}{blue} blocks represent the memory and current chunk, respectively. The number in each block represents the temporal index in RoPE. For simplicity of illustration, each chunk only contains one frame.}
  \label{fig:attention_comp}
\end{figure}

\tf{Maintaining long-term geometric consistency requires recalling past frames, ensuring content remains unchanged when revisiting to a previous location. However, naively using all past frames as context (Fig.~\ref{fig:attn-a}) is computationally intractable and redundant for long sequences.  To address this, we rebuild a memory context $C_{t}$  from past chunks $O_{t-1}$ for each new chunk $x_t$. Our approach advances beyond prior work~\cite{xiao2025worldmem, yu2025context, chen2025learning} by combining both short-term temporal cues and long-range spatial references: 1) A temporal memory ($C_{t}^{T}$) comprises $L$ most recent chunks $\{x_{t-L}, ..., x_{t-1}\}$  to ensure short-term motion smoothness.  2) A spatial memory ($C_{t}^{S}$) samples from non-adjacent past frames to prevent geometric drift over long sequences, where $C_{t}^{S} \subseteq O_{t-1}-C_{t}^{T}$. This sampling is guided by geometric relevance scores that incorporate both FOV overlap and camera distance.
}

\tf{Once memory context is rebuilt, the challenge shifts to applying them to enforce consistency. Effectively using retrieved context requires overcoming a fundamental flaw in positional encodings. With standard RoPE (Fig.\ref{fig:attn-b}), the distance between the current chunk and past memory grows unbounded over time. This growing relative distance  can eventually exceed the trained interpolation range in RoPE, causing extrapolation artifacts~\cite{su2024roformer}. More critically, the growing perceived distance to these long-past spatial memory would weaken their influence on the current prediction. To resolve this, we propose Temporal Reframing (Fig.\ref{fig:attn-c}).  We discard the absolute temporal indices, and dynamically re-assign new positional encodings to all context frames, establishing a fixed, small relative distance to the current, irrespective of their actual temporal gap.
This operation effectively ``pulls" important past frames closer in time, ensuring their sustained influence and enabling robust extrapolation for long-term consistency.
}

\subsection{Context Forcing}
\label{sec:distill}

Autoregressive models often suffer from error accumulation during long video generation, leading to degraded visual quality over time~\cite{huang2025self, yin2025slow}. Moreover, the multi-step denoising of diffusion models is too slow for real-time interaction. 
\tf{Recent methods~\cite{huang2025self, yang2025longlive, liu2025rolling, cui2025self} address these challenges by distilling a powerful bidirectional teacher diffusion model into a fast, few-step autoregressive student.  These techniques force the student's output distribution $p_{\theta}(x_{0:t})$ to  align with the teacher's, thereby improving generation quality by employing a distribution matching loss~\cite{yin2024improved}:}
\begin{equation}
     \nabla_{\theta} \mathcal{L}_{DMD}=\mathbb{E}_{k}(\nabla_{\theta}\text{KL}(p_{\theta}(x_{0:t})||p_{data}(x_{0:t}))),
\end{equation}
\noindent where the gradient of the reverse KL can be approximated by the score difference derived from teacher model.

\begin{figure}
  \centering
  \includegraphics[width=1\linewidth]{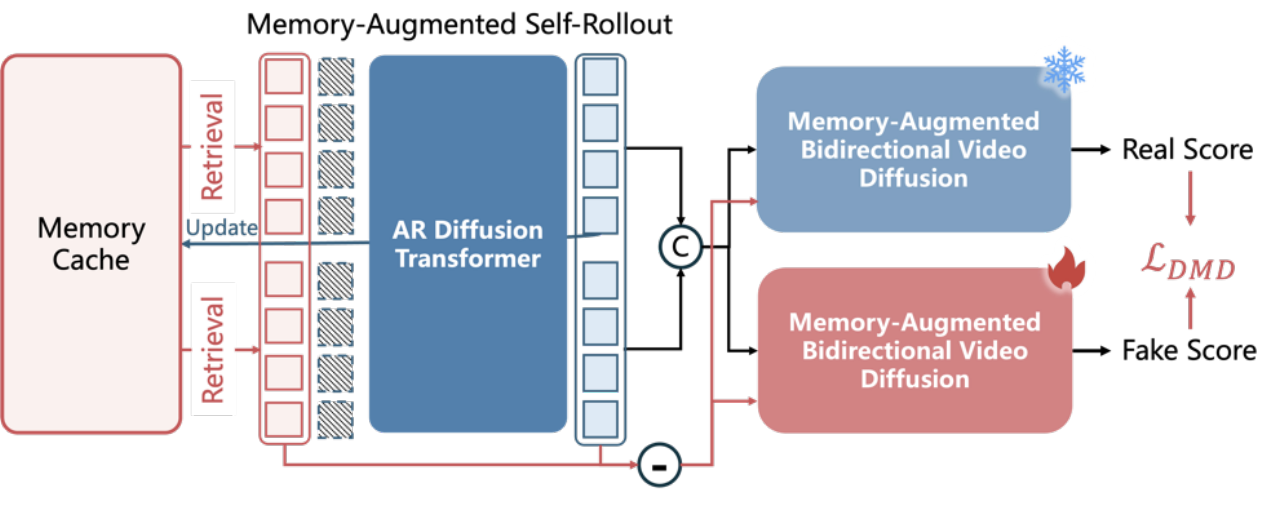}
    \vspace{-8mm}
  \caption{\textbf{Context forcing} is a novel distillation method that employs memory-augmented self-rollout and memory-augmented bidirectional video diffusion to preserve long-term consistency, enable real-time interaction, and mitigate error accumulation.
  }
  \label{fig:distillation}
  \vspace{-4mm}
\end{figure}

\begin{table*}[t]

\caption{
\textbf{Quantitative comparisons.} We compare against both methods without memory, \ie, CameraCtrl~\cite{he2024cameractrl}, SEVA~\cite{zhou2025stable}, ViewCrafter~\cite{yu2024viewcrafter}, Matrix-Game-2.0~\cite{he2025matrix}, and GameCraft~\cite{li2025hunyuan}, and methods with memory, \ie, Gen3C~\cite{ren2025gen3c}, VMem~\cite{li2025vmem}. Our method achieves superior results, particularly in long-term settings, which more clearly demonstrate the long-term consistency.
}
\label{tab:long-term}
\vspace{-1mm}
\centering
\scriptsize
\renewcommand{\arraystretch}{1.1}
\newcommand{\gou}{\textcolor{ForestGreen}{\ding{52}}}
\newcommand{\cha}{\textcolor{Red}{\ding{55}}}
\begin{tabular}{lcccccc|ccccc}
\toprule
\multicolumn{2}{c}{} &
\multicolumn{5}{c}{\textbf{Short-term} ($61 $ \text{frames})} &
\multicolumn{5}{c}{\textbf{Long-term} ($\geq 250 $ \text{frames})} \\
\cmidrule(lr){3-7}\cmidrule(lr){8-12}
\multicolumn{1}{c}{} &
\multicolumn{1}{c}{\textbf{Real-time}} &
\textbf{PSNR} $\uparrow$ &
\textbf{SSIM} $\uparrow$ &
\textbf{LPIPS} $\downarrow$ &
$R_{\text{dist}}$ $\downarrow$ &
$T_{\text{dist}}$ $\downarrow$ &
\textbf{PSNR} $\uparrow$ &
\textbf{SSIM} $\uparrow$ &
\textbf{LPIPS} $\downarrow$ &
$R_{\text{dist}}$ $\downarrow$ &
$T_{\text{dist}}$ $\downarrow$ \\
\midrule
   CameraCtrl~\cite{he2024cameractrl}
    & \cha & 17.93 & 0.569  & 0.298 & 0.037 & 0.341
    & 10.09 & 0.241  & 0.549  & 0.733 & 1.117 \\

   SEVA~\cite{zhou2025stable}
    & \cha & 19.84 & 0.598   & 0.313 & 0.047 & 0.223 
    & 10.51 & 0.301  & 0.517 & 0.721 & 1.893  \\
    
   ViewCrafter~\cite{yu2024viewcrafter}
    & \cha  & 19.91 & 0.617  & 0.327 & 0.029 & 0.543
    & 9.32 & 0.277  & 0.661 & 1.573 & 3.051 \\
   Gen3C~\cite{ren2025gen3c}
    & \cha &  21.68 & 0.635  & 0.278  & \textbf{0.024} & 0.477
    & 15.37 &  0.431 & 0.483  & {0.357} & 0.979 \\

   VMem~\cite{wang2024motionctrl}
    & \cha & 19.97  & 0.587   & 0.316  & 0.048 & 0.219
    & 12.77 & 0.335  & 0.542  & 0.748 & 1.547 \\

    Matrix-Game-2.0~\cite{he2025matrix}
    & \gou & 17.26 & 0.505  & 0.383   & 0.287 & 0.843
    & 9.57 & 0.205  & 0.631  & 2.125 & 2.742 \\

    GameCraft~\cite{li2025hunyuan}
    & \cha &  21.05 & 0.639  & 0.341  & 0.151 & 0.617
    & 10.09 & 0.287  & 0.614  & 2.497 & 3.291 \\

    \midrule

    Ours (w/o Context Forcing)
    & \cha & 21.27 & 0.669  & 0.261   & 0.033 & 0.157
    & 16.27 & 0.425 & 0.495 & 0.611 & 0.991  \\

    Ours (full)
    & \gou & \textbf{21.92} & \textbf{0.702} & \textbf{0.247}  & 0.031 & \textbf{0.121}
    & \textbf{18.94} & \textbf{0.585} &  \textbf{0.371}  & \textbf{0.332} & \textbf{0.797}  \\
\bottomrule
\end{tabular}
\vspace{-0mm}

\vspace{-4mm}
\end{table*}

\tf{However, these methods are incompatible with memory-aware models due to a critical distribution mismatch. Standard teacher diffusion models are trained on short clips and are inherently memory-less. Even if a teacher is augmented with memory, its  bidirectional nature inevitably differs from the student's causal, autoregressive process. This means that without a meticulously designed memory context to mitigate this gap,  the difference in memory context will make their conditional distributions $p(x|C)$ misaligned, which in turn causes distribution matching to fail.}

\tf{We thus propose context forcing as shown in Fig.~\ref{fig:distillation}, which alleviates  the memory context misalignment between  teacher and  student for distillation. For the student model, 
we self-rollout 4 chunks conditioned on the memory context $p_{\theta}(x_{j:j+3}|x_{0:j-1})=\prod \limits_{i=j}^{j+3} p_{\theta}(x_{i}|C_{i})$.}
\tf{To construct our teacher model $V_{\beta}$, we augment a standard bidirectional diffusion model with memory, and structure its context  by masking $x_{j:j+3}$ from student's memory context, }
\begin{equation}
     p_{data}(x_{j:j+3}|x_{0:j-1})=p_{\beta}(x_{j:j+3}|C_{j:j+3}-x_{j:j+3}),
\end{equation}
where $C_{j:j+3}$ denotes all context memory chunks corresponding to student's self-rollout $x_{j:j+3}$. By aligning the memory context with the student model, we enforce the distributions represented by the teacher to be as close as possible to the student model, which enables more effectively distribution matching. Moreover, this avoids training $V_{\beta}$ on long videos and redundant context, facilitating the learning of long-term visual distribution. Additionally, we introduce a progressive distillation strategy that incrementally increases the number of self-rollout latents. This facilitates distillation across varying sequence lengths, thereby enhancing long-horizon video generation. Through context forcing, we preserve long-term consistency in real-time generation with 4-denoising steps, and mitigate error accumulation.

\begin{figure*}[t]
  \centering
  \includegraphics[width=1\linewidth]{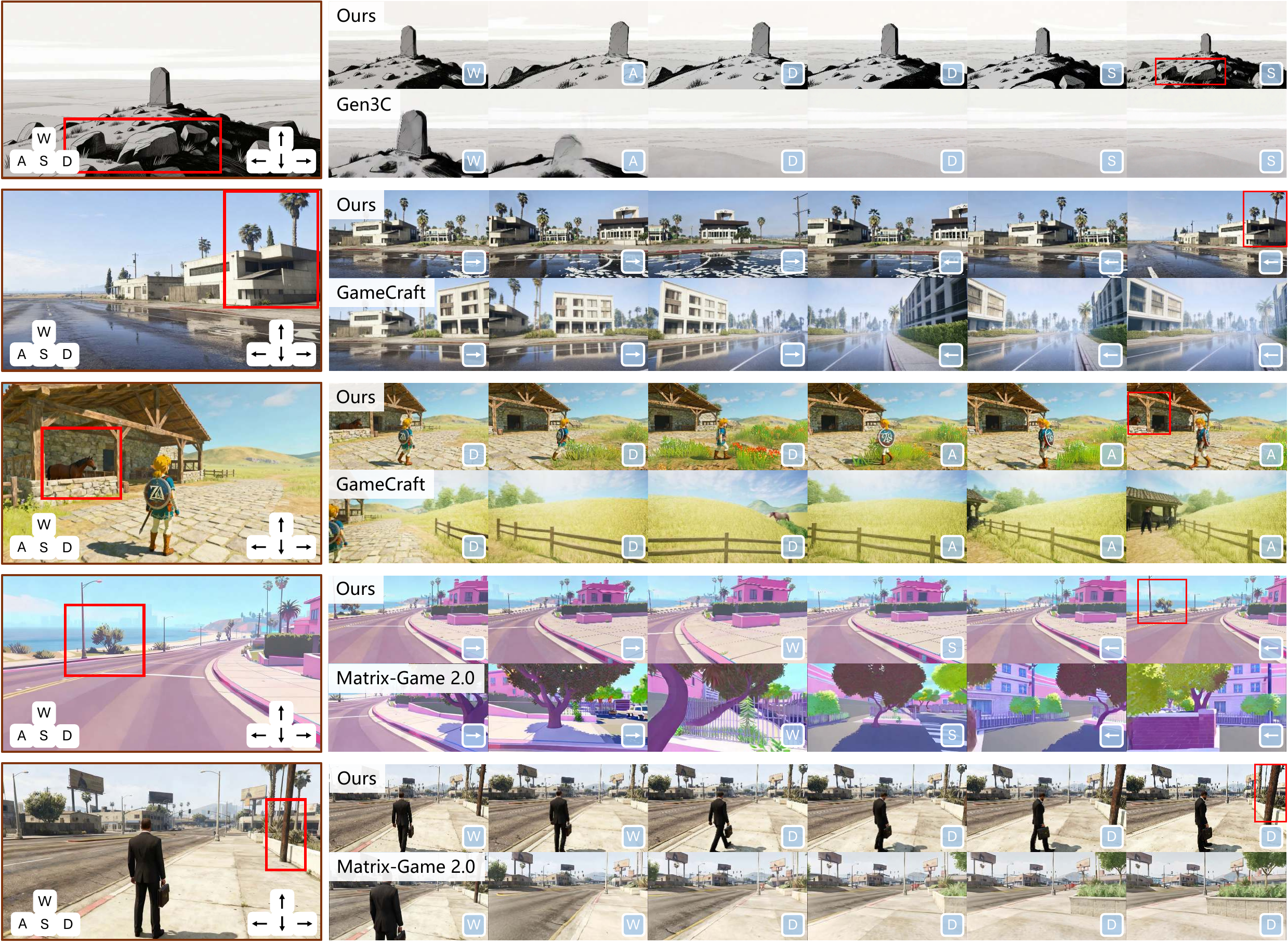}
    \vspace{-6mm}
   \caption{\textbf{Qualitative comparisons with existing methods.} WorldPlay achieves the state-of-the-art long-term consistency (\textit{shown in red boxes}) and visual quality across diverse scenes, including both first- and third-person real and stylized worlds.}
   \label{fig:main_res}
   \vspace{-2mm}
\end{figure*}

\subsection{Streaming Generation with Real-Time Latency}
\label{sec:stream}
\tf{We augment context forcing with a suite of optimizations to minimize latency, unlocking an interactive streaming experience at 24 FPS and 720p resolution on 8×H800 GPUs.}

\noindent\textbf{Mixed Parallelism Method for DiT and VAE.} 
Unlike the conventional parallelism method that replicates the entire model or adapting sequence parallelism on the temporal dimension, our parallelism method combines sequence parallelism~\cite{li-etal-2023-sequence} and attention parallelism, which partitions the tokens of each chunk across devices. 
This design ensures that the computational workload is distributed evenly, substantially reducing per-chunk inference time while maintaining generation quality.

\noindent\textbf{Streaming Deployment and Progressive Decoding.}
To minimize time-to-first-frame and enable seamless interaction, we adopt a streaming deployment architecture using NVIDIA Triton Inference Framework 
and implement a progressive VAE decoding strategy that decodes and streams frames in smaller batches, 
allowing users to observe generated content while subsequent frames are still being processed. This streaming pipeline 
ensures smooth, low-latency interaction even under varying computational loads.

\noindent\textbf{Quantization and Efficient Attention.}
Furthermore, we employ a comprehensive suite of quantization strategies. Specifically, we adopt Sage Attention~\cite{zhang2025sageattention}, float quantization, and matrix multiplication quantization to improve the inference performance. Additionally, we use KV-cache mechanisms for attention modules to eliminate redundant computations during autoregressive generation. 

\section{Experiments}

\noindent \textbf{Implementation Details.} WorldPlay is trained on a comprehensive dataset comprising approximately 320K high-quality video samples derived from both real-world footage and synthetic environments. Details regarding the dataset processing pipeline and the training/inference are provided in Appendix~\ref{supp_B} and Appendix~\ref{supp_A}, respectively.

\noindent \textbf{Evaluation Protocol.} 
\tf{Our test set comprises 600 cases sourced from DL3DV, game videos, and AI-generated images spanning a range of styles.}
For the short-term setting, we utilize the camera trajectories from the test videos as the input pose. The generated frames are directly compared against the ground truth to assess visual quality and action precision. For the long-term setting, we test the long-horizon generation ability and long-term consistency using various custom cycle camera trajectories designed to enforce revisiting. Each model generates frames along a customize trajectory and then returns along the same path, metrics are evaluated on the return path by comparing the generated frame to the corresponding frame generated during the initial pass. We employ LPIPS, PSNR, and SSIM to measure visual quality and $R_{\text{dist}}$ and $T_{\text{dist}}$ to quantify action precision. 

\wenq{
Specifically, we use the ground truth poses as the input for each model to generate videos. Then, we utilize ViPE to estimate the camera pose of the generated videos. Following the previous works~\cite{he2024cameractrl,yu2024viewcrafter}, we compute the relative poses of the ground truth and generated camera poses by setting the extrinsic matrix of first frame as an identity matrix and normalize the translation scale using the furthest frame. The rotation distance $R_{\text{dis}}$ is calculated by comparing the predicted rotation matrices $R_{gen}$ and ground truth rotation matrices $R_{gt}$:
\begin{align}
R_{\text{dis}} = \arccos \left( \frac{\text{tr}(\mathbf{R}_{\text{gen}} \mathbf{R}_{\text{gt}}^{\top}) - 1}{2} \right),
\end{align}
where $\text{tr}(\cdot)$ denotes the trace of the matrix. The translation distance $T_{dis}$ is computed between the predicted $\mathbf{t}_{\text{gen}}$ and ground truth translation vectors $\mathbf{t}_{\text{gt}}$:
\begin{align}
T_{\text{dis}} = \| \mathbf{t}_{\text{gt}} - \mathbf{t}_{\text{gen}} \|^2_2.
\end{align}
}

\vspace{-6mm}
\noindent \textbf{Baselines.} We conduct comprehensive comparisons against various baselines, which mainly fall into two categories: 1) \textit{Action-controlled diffusion models without memory}:  CameraCtrl~\cite{he2024cameractrl}, SEVA~\cite{zhou2025stable}, ViewCrafter~\cite{yu2024viewcrafter}, Matrix-Game 2.0~\cite{he2025matrix} and GameCraft~\cite{li2025hunyuan}; 2) \textit{Action-controlled diffusion models with memory}: Gen3C~\cite{ren2025gen3c} and VMem~\cite{li2025vmem}. More evaluation results can be found in Appendix.

\subsection{Main Result} \label{main_res}

\noindent \textbf{Quantitative Results.} As shown in Table~\ref{tab:long-term}, in the short-term regime, our approach achieves superior visual fidelity and maintains competitive control accuracy. Although methods leveraging explicit 3D representations (\ie ViewCrafter~\cite{yu2024viewcrafter}, Gen3C~\cite{ren2025gen3c}) realize more accurate rotation, they suffer from issues such as the inaccurate depth estimation and inconsistent scale when performing translations. For more challenging long-term scenarios, where action accuracy generally degrades, our method remains more stable and achieves the best performance. Regarding long-term geometric consistency, Matrix-Game-2.0~\cite{he2025matrix} and GameCraft~\cite{li2025hunyuan} exhibit poor performance due to the lack of memory mechanism. Although VMem~\cite{li2025vmem} and Gen3C~\cite{ren2025gen3c} employ explicit 3D cache to maintain consistency, they are constrained by depth accuracy and alignment, making it difficult to achieve robust long-term consistency. Benefiting from Reconstituted Context Memory, we achieve improved long-term consistency. Moreover, through context forcing, we further prevent error accumulation, resulting in better visual quality and action accuracy. \wenq{To rigorously evaluate the long-horizon 3D structural consistency of our model, we incorporate a more advanced metric, MEt3R \cite{asim2025met3r}, which explicitly models the multi-view correspondence by leveraging pre-trained 3D geometric priors (e.g., DUSt3R \cite{Wang_2024_CVPR}). We evaluate the generated long-horizon videos by pairing frames from the initial trajectory with the corresponding frames in the revisiting trajectory, which are then fed into the MEt3R model to quantify their 3D structural alignment. As shown in Table.~\ref{tab:met3r_results}, it clearly demonstrates the superiority of our approach in maintaining precise 3D consistency.}

To further validate the efficiency of our optimization strategies proposed in Sec. \ref{sec:stream}, we conduct experiments on inference speed as detailed in Table~\ref{tab:inference_speed_transformed}, demonstrating that the tailored parallelization and quantization applied to DiT and VAE significantly boost the inference throughput of our model. Crucially, WorldPlay concurrently achieves the requisite real-time interactivity for immersive simulation.

\begin{table}[t]
\centering
\caption{\textbf{Quantitative Comparison of 3D Structural Consistency} using MEt3R~\cite{asim2025met3r}.}
\label{tab:met3r_results}
\scriptsize 
\begin{tabular}{lc}
\toprule
\textbf{Method} & \textbf{MEt3R Score ($\downarrow$)} \\
\midrule
Gen3C~\cite{ren2025gen3c}           & 0.187 \\
Matrix-Game-2.0~\cite{he2025matrix}     & 0.367 \\
GameCraft~\cite{li2025hunyuan}       & 0.305 \\
\textbf{Ours (full)}   & \textbf{0.133} \\
\bottomrule
\vspace{-4mm}
\end{tabular}
\end{table}

\begin{table}[t]
\centering
\caption{\textbf{Quantitative evaluation of inference acceleration strategies.} "P" and "Q" denote Parallelism and Quantization, respectively.}
\label{tab:inference_speed_transformed}
\scriptsize 
\begin{tabular}{ccc|cc}
\toprule
\textbf{DiT P\&Q} & \textbf{VAE P\&Q} & \textbf{Streaming Decoding} & \textbf{FPS} & \textbf{Improvement} \\
\midrule
& & & 2.80 & --- \\
\checkmark & & & 3.80 & $1.35\times$ \\
\checkmark & \checkmark & & 16.0 & $5.71\times$ \\
\checkmark & \checkmark & \checkmark & 24.0 & $8.57\times$ \\
\bottomrule
\end{tabular}
\end{table}

\begin{table}[t]
\caption{
\textbf{Ablation for action representation.}  }
\vspace{-2mm}
\label{tab:action}
\centering
\scriptsize 
\begin{tabular}{l|cccccc}
\toprule
\textbf{Action} & \textbf{PSNR$\uparrow$} & \textbf{SSIM$\uparrow$} & \textbf{LPIPS$\downarrow$} & $R_{\text{dist}}$ $\downarrow$ & $T_{\text{dist}}$ $\downarrow$ \\
\midrule
Discrete  & 21.47 & 0.661 & 0.248 & 0.103 &  0.615 \\
Continuous  & 21.93 & 0.665 & 0.231 & 0.038 & 0.287 \\
Full  & \textbf{22.09} & \textbf{0.687} & \textbf{0.219} & \textbf{0.028} & \textbf{0.113} \\
\bottomrule
\end{tabular}
\end{table}

\noindent \textbf{Qualitative Results.} We provide qualitative comparisons with baselines in Fig. \ref{fig:main_res}. The explicit 3D cache used in Gen3C~\cite{ren2025gen3c} is highly sensitive to the quality of intermediate output and limited by the accuracy of depth estimation. Conversely, our reconstituted context memory guarantees long-term consistency with more robust implicit prior, achieving superior generalizability. Matrix-Game-2.0~\cite{he2025matrix} and GameCraft~\cite{li2025hunyuan} fail to support free exploration due to the lack of memory. Furthermore, they do not generalize well to third-person scenarios, making it difficult to control agents and limiting their applicability. In contrast, WorldPlay successfully extends its efficacy to these scenarios and maintains high visual fidelity and long-term geometric consistency.

\subsection{Ablation} \label{ablation}

\textbf{Action Representation.} Table~\ref{tab:action} validates the effectiveness of the proposed dual-action representation. When using only discrete keys as action signals, the model struggles to achieve fine-grained control, such as the distance of movement or the degree of rotation, resulting in poor performance on $R_{\text{dist}}$ and $T_{\text{dist}}$ metrics. Using continuous camera poses yields better results but converges more difficult due to scale variance. By employing the dual-action representation, we achieve the best overall control performance.

\begin{table}[t]
\caption{
\textbf{Ablation for positional encoding design in memory.} The results are evaluated on the long-term test data.
}
\vspace{-2mm}
\label{tab:rope}
\centering
\scriptsize 
\begin{tabular}{l|ccccc}
\toprule
\textbf{} & \textbf{PSNR$\uparrow$} & \textbf{SSIM$\uparrow$} & \textbf{LPIPS$\downarrow$} & $R_{\text{dist}}$ $\downarrow$ & $T_{\text{dist}}$ $\downarrow$ \\
\midrule
RoPE  & 14.03 & 0.358 & 0.534 & 0.805  & 1.341 \\
  Reframed RoPE  & \textbf{16.27} & \textbf{0.425} & \textbf{0.495} & \textbf{0.611} & \textbf{0.991} \\
\bottomrule
\end{tabular}
\vspace{-3mm}
\end{table}

\begin{figure}[t]
  \centering
  \includegraphics[width=1\linewidth]{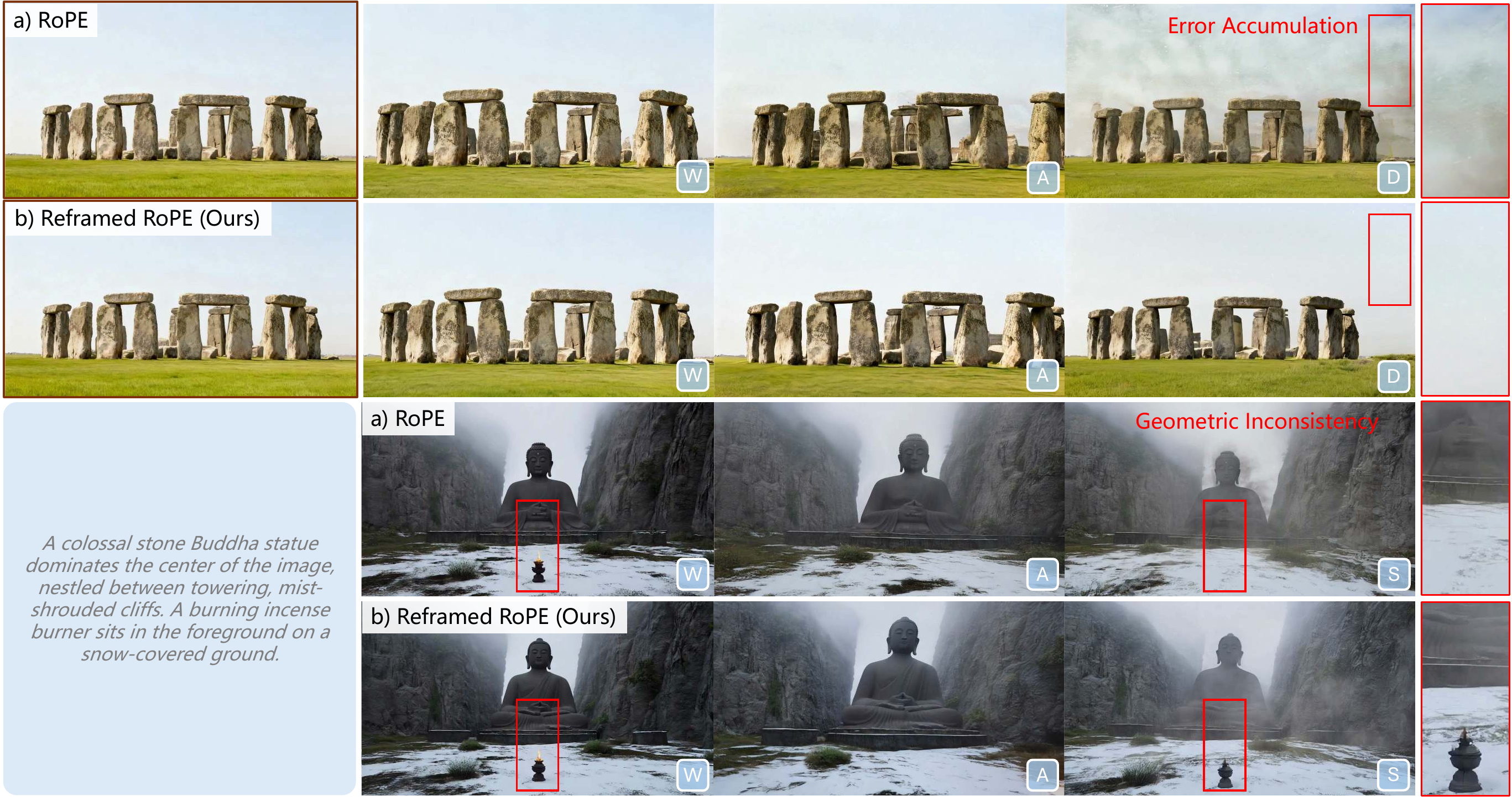}
  \caption{\textbf{RoPE design comparisons. Upper:} Our reframed RoPE avoids exceeding the the positional range in standard RoPE, alleviating error accumulation. \textbf{Bottom:} By maintaining a small relative distance to long-range spatial memory, it achieves better long-term consistency.
  }
  \label{fig:rope}
\end{figure}

\noindent \textbf{RoPE Design.} Table.~\ref{tab:rope} presents the quantitative results of different RoPE designs within the memory mechanism \wenq{as detailed in Sec.~\ref{sec:memory}}, showing that reframed rope outperforms naive counterparts, especially on visual metrics. As illustrated in the upper part of Fig.~\ref{fig:rope}, RoPE is more prone to error accumulation. It also increases the distance between memory and predicted chunk due to absolute temporal indices, resulting in weaker geometric consistency, as shown in the lower part of Fig.~\ref{fig:rope}.

\begin{figure}[t]
  \centering
  \includegraphics[width=1\linewidth]{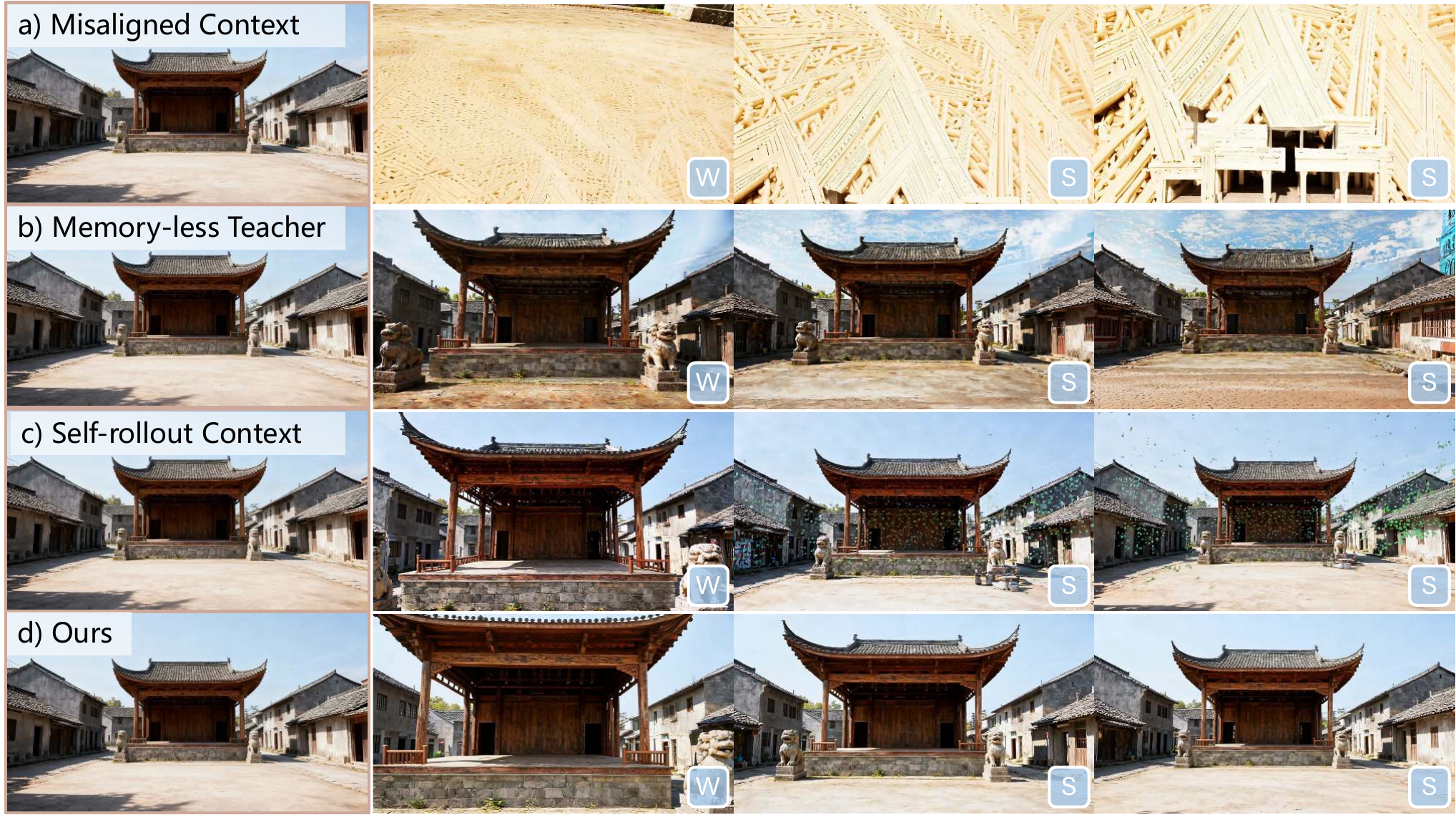}
    \vspace{-5mm}
  \caption{\textbf{Ablation for context forcing.} \textbf{a)} When the teacher and student have misaligned context, it leads to distillation failure, resulting in collapsed outputs. \textbf{b)} Leveraging memory-less teacher introduces a distribution mismatching. \textbf{c)} Self-rollout historical context can introduce artifacts. Zoom in for details.
  }
  \label{fig:distillation_ablation}
\end{figure}

\begin{figure}[t]
  \centering
  \includegraphics[width=1\linewidth]{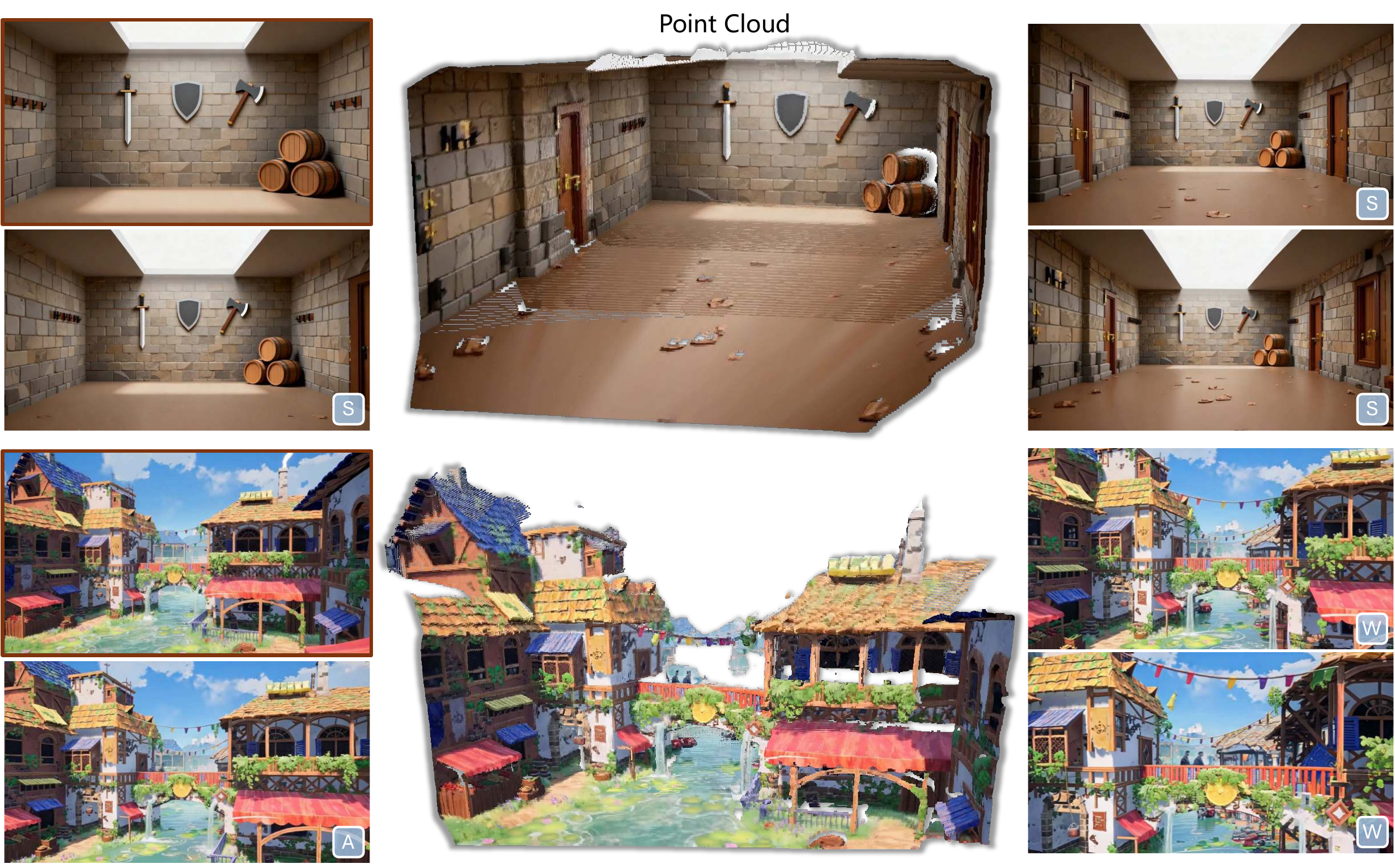}
  \caption{\textbf{3D reconstruction results.} We first utilize our model to autoregressively generate videos. The videos are then processed by a 3D reconstruction model to produce the final point clouds.
  }
  \label{fig:supp_3d_recon}
\end{figure}

\noindent \textbf{Context Forcing.} To verify the importance of memory alignment, we train the teacher model following~\cite{yu2025context}, where the memory is selected at latent level rather than at chunk level. Although this may reduce the number of memory context in the teacher model, it also introduce misaligned context between the teacher and student model, leading to collapsed results as shown in Fig.~\ref{fig:distillation_ablation}a. Moreover, utilizing a memory-less bidirectional model as the teacher induces a distribution mismatch, which hinders long-horizon video generation and significantly compromises long-term consistency, as illustrated in Fig.~\ref{fig:distillation_ablation}b. Additionally, for the past chunks $x_{0:j-1}$, we attempt to self-rollout historical chunks as context following~\cite{yang2025longlive}. However, this may cause the bidirectional diffusion model to provide inaccurate score estimation, as it is trained using clean chunks as memory. Consequently, this discrepancy introduces artifacts as illustrated in~Fig.~\ref{fig:distillation_ablation}c. We obtain historical chunks by sampling from real videos, which yields superior results as shown in Fig.~\ref{fig:distillation_ablation}d.

\subsection{Application} \label{application}

\noindent \textbf{3D Reconstruction.} Benefiting from the long-term  geometric consistency, we can integrate a feed-forward 3D reconstruction model~\cite{liu2025worldmirror} to produce high-quality point clouds \wenq{from the generated videos}, as presented in Fig. \ref{fig:teaser} (d) and Fig.~\ref{fig:supp_3d_recon}.

\begin{figure}[t]
  \centering
  \includegraphics[width=1\linewidth]{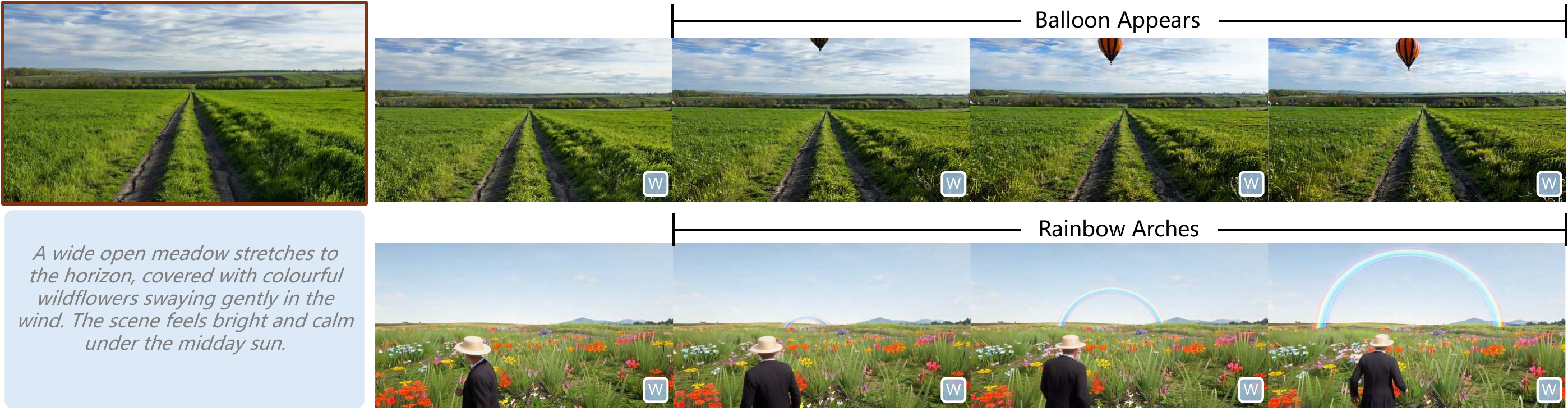}
    \vspace{-5mm}
  \caption{\textbf{Promptable event.} Our method supports text-based manipulation during streaming.
  }
  \label{fig:event}
  \vspace{-6mm}
\end{figure}

\noindent \textbf{Promptable Event.} 
\tf{Beyond navigation control, WorldPlay supports text-based interaction to trigger dynamic world events \wenq{(\ie, environmental transitions and object appearances)}. As shown in Fig.~\ref{fig:event} and Fig.~\ref{fig:teaser} (e), users can prompt at any time  to responsively alter the ongoing stream.}
 
\section{Conclusion}

\tf{WorldPlay is a powerful world model with real-time interaction and long-term geometric consistency. It empowers users to customize unique worlds from a single image or text prompt. While focused on navigation control,  its architecture has shown potential for richer interaction like dynamic, text-triggered events.   By providing a systematic framework for control, memory, and distillation, WorldPlay marks a critical step toward creating consistent and interactive virtual worlds. }

\wenq{
\textbf{Limitations.} While WorldPlay demonstrates strong performance, several avenues remain open for exploration and improvement. First, our model can generawting videos of approximately 30 seconds, efficiently scaling this framework to longer durations, such as minutes or hours~\cite{cui2025self,cui2026lol}, remains a significant challenge. Second, although distillation is utilized to mitigate error accumulation, fundamentally averting this phenomenon during the training of autoregressive diffusion models remains a critical open challenge. Moreover, expanding the action types to a broader set with multi-agent interaction and complex physical dynamics is another promising direction. Finally, retrieval mechanisms based on the FOV may fail to accurately identify memory context when faced with significant occlusions~\cite{yu2025context}. 
}

\section*{Acknowledgement} 
This work was supported by the Hong Kong Research Grants Council under the Areas of Excellence scheme grant AoE/E-601/22-R and NSFC/RGC Collaborative Research Scheme grant CRS$\_$HKUST603/22.

\section*{Impact Statement}

This paper presents work whose goal is to advance the field of Machine
Learning. There are many potential societal consequences of our work, none which we feel must be specifically highlighted here.




\nocite{langley00}

\bibliography{example_paper}

@misc{oasis,
  author = {Decart, Etched},
  title = {Oasis: A Universe in a Transformer},
  howpublished = {\url{https://oasis-model.github.io/}},
  year = {2024}, 
}

@article{he2025matrix,
  title={{Matrix-Game} 2.0: An open-source, real-time, and streaming interactive world model},
  author={He, Xianglong and Peng, Chunli and Liu, Zexiang and Wang, Boyang and Zhang, Yifan and Cui, Qi and Kang, Fei and Jiang, Biao and An, Mengyin and Ren, Yangyang and others},
  journal={arXiv preprint arXiv:2508.13009},
  year={2025}
}

@article{li2025hunyuan,
  title={{Hunyuan-GameCraft}: High-dynamic Interactive Game Video Generation with Hybrid History Condition},
  author={Li, Jiaqi and Tang, Junshu and Xu, Zhiyong and Wu, Longhuang and Zhou, Yuan and Shao, Shuai and Yu, Tianbao and Cao, Zhiguo and Lu, Qinglin},
  journal={arXiv preprint arXiv:2506.17201},
  year={2025}
}

@InProceedings{ho2020denoising,
  title={Denoising diffusion probabilistic models},
  author={Ho, Jonathan and Jain, Ajay and Abbeel, Pieter},
  booktitle={Advances in Neural Information Processing Systems},
  volume={33},
  pages={6840--6851},
  year={2020}
}

@inproceedings{lipman2022flow,
    author = {Lipman, Yaron and Chen, Ricky TQ and Ben-Hamu, Heli and Nickel, Maximilian and Le, Matt},
    title = {Flow matching for generative modeling},
    booktitle = {ICLR},
    year = 2023
}

@inproceedings{chen2024videocrafter2,
  title={{VideoCrafter2}: Overcoming data limitations for high-quality video diffusion models},
  author={Chen, Haoxin and Zhang, Yong and Cun, Xiaodong and Xia, Menghan and Wang, Xintao and Weng, Chao and Shan, Ying},
  booktitle = {CVPR},
  pages={7310--7320},
  year={2024}
}

@inproceedings{guo2023animatediff,
  title={{AnimateDiff}: Animate your personalized text-to-image diffusion models without specific tuning},
  author={Guo, Yuwei and Yang, Ceyuan and Rao, Anyi and Liang, Zhengyang and Wang, Yaohui and Qiao, Yu and Agrawala, Maneesh and Lin, Dahua and Dai, Bo},
  booktitle = {ICLR},
  year = 2024
}

@inproceedings{yang2024cogvideox,
  title={{CogVideoX}: Text-to-video diffusion models with an expert transformer},
  author={Yang, Zhuoyi and Teng, Jiayan and Zheng, Wendi and Ding, Ming and Huang, Shiyu and Xu, Jiazheng and Yang, Yuanming and Hong, Wenyi and Zhang, Xiaohan and Feng, Guanyu and others},
  booktitle = {ICLR},
  year = 2024
}

@inproceedings{rombach2022high,
  title={High-resolution image synthesis with latent diffusion models},
  author={Rombach, Robin and Blattmann, Andreas and Lorenz, Dominik and Esser, Patrick and Ommer, Bj{\"o}rn},
  booktitle={CVPR},
  pages={10684--10695},
  year={2022}
}

@inproceedings{peebles2023scalable,
  title={Scalable diffusion models with transformers},
  author={Peebles, William and Xie, Saining},
  booktitle={ICCV},
  pages={4195--4205},
  year={2023}
}

@misc{veo,
 author = {Google Deepmind},
 title = {Veo3 video model},
 note = {\url{https://deepmind.google/models/veo/}},
 year = 2025
}

@article{wan2025wan,
  title={Wan: Open and advanced large-scale video generative models},
  author={Wan, Team and Wang, Ang and Ai, Baole and Wen, Bin and Mao, Chaojie and Xie, Chen-Wei and Chen, Di and Yu, Feiwu and Zhao, Haiming and Yang, Jianxiao and others},
  journal={arXiv preprint arXiv:2503.20314},
  year={2025}
}

@misc{kling,
 author = {Kuaishou},
 title = {Kling video model},
 note = {\url{https://klingai.com/global/}},
 year = 2024
}

@misc{hailuo,
 author = {Minimax},
 title = {Hailuo video model},
 note = {\url{https://hailuoai.video}},
 year = 2024
}

@article{kong2024hunyuanvideo,
  title={{HunyuanVideo}: A systematic framework for large video generative models},
  author={Kong, Weijie and Tian, Qi and Zhang, Zijian and Min, Rox and Dai, Zuozhuo and Zhou, Jin and Xiong, Jiangfeng and Li, Xin and Wu, Bo and Zhang, Jianwei and others},
  journal={arXiv preprint arXiv:2412.03603},
  year={2024}
}

@article{gao2025seedance,
  title={Seedance 1.0: Exploring the Boundaries of Video Generation Models},
  author={Gao, Yu and Guo, Haoyuan and Hoang, Tuyen and Huang, Weilin and Jiang, Lu and Kong, Fangyuan and Li, Huixia and Li, Jiashi and Li, Liang and Li, Xiaojie and others},
  journal={arXiv preprint arXiv:2506.09113},
  year={2025}
}

@article{wiedemer2025video,
  title={Video models are zero-shot learners and reasoners},
  author={Wiedemer, Thadd{\"a}us and Li, Yuxuan and Vicol, Paul and Gu, Shixiang Shane and Matarese, Nick and Swersky, Kevin and Kim, Been and Jaini, Priyank and Geirhos, Robert},
  journal={arXiv preprint arXiv:2509.20328},
  year={2025}
}

@inproceedings{song2020score,
  title={Score-based generative modeling through stochastic differential equations},
  author={Song, Yang and Sohl-Dickstein, Jascha and Kingma, Diederik P and Kumar, Abhishek and Ermon, Stefano and Poole, Ben},
  booktitle = {ICLR},
  year = 2021
}

@inproceedings{chen2024diffusion,
  title={{Diffusion Forcing}: Next-token prediction meets full-sequence diffusion},
  author={Chen, Boyuan and Mart{\'\i} Mons{\'o}, Diego and Du, Yilun and Simchowitz, Max and Tedrake, Russ and Sitzmann, Vincent},
  booktitle={Advances in Neural Information Processing Systems},
  volume={37},
  pages={24081--24125},
  year={2024}
}

@inproceedings{henschel2025streamingt2v,
  title={{StreamingT2V}: Consistent, dynamic, and extendable long video generation from text},
  author={Henschel, Roberto and Khachatryan, Levon and Poghosyan, Hayk and Hayrapetyan, Daniil and Tadevosyan, Vahram and Wang, Zhangyang and Navasardyan, Shant and Shi, Humphrey},
  booktitle={CVPR},
  pages={2568--2577},
  year={2025}
}

@inproceedings{kim2024fifo,
  title={{FIFO-Diffusion}: Generating infinite videos from text without training},
  author={Kim, Jihwan and Kang, Junoh and Choi, Jinyoung and Han, Bohyung},
  booktitle={Advances in Neural Information Processing Systems},
  volume={37},
  pages={89834--89868},
  year={2024}
}

@inproceedings{alonso2024diffusion,
  title={Diffusion for world modeling: Visual details matter in {Atari}},
  author={Alonso, Eloi and Jelley, Adam and Micheli, Vincent and Kanervisto, Anssi and Storkey, Amos J and Pearce, Tim and Fleuret, Fran{\c{c}}ois},
  booktitle={Advances in Neural Information Processing Systems},
  volume={37},
  pages={58757--58791},
  year={2024}
}

@article{parkerholder2024genie2,
  title         = {Genie 2: A Large-Scale Foundation World Model},
  author        = {Jack Parker-Holder and Philip Ball and Jake Bruce and Vibhavari Dasagi and Kristian Holsheimer and Christos Kaplanis and Alexandre Moufarek and Guy Scully and others},
  year          = {2024},
  url           = {https://deepmind.google/discover/blog/genie-2-a-large-scale-foundation-world-model/}
}

@inproceedings{bar2025navigation,
  title={Navigation world models},
  author={Bar, Amir and Zhou, Gaoyue and Tran, Danny and Darrell, Trevor and LeCun, Yann},
  booktitle={CVPR},
  pages={15791--15801},
  year={2025}
}

@inproceedings{valevski2024diffusion,
  title={Diffusion models are real-time game engines},
  author={Valevski, Dani and Leviathan, Yaniv and Arar, Moab and Fruchter, Shlomi},
  booktitle = {ICLR},
  year = 2025
}

@inproceedings{li2025flash,
  title={{FlashWorld}: High-quality {3D} Scene Generation within Seconds},
  author={Xinyang Li and Tengfei Wang and Zixiao Gu and Shengchuan Zhang  and Chunchao Guo and Liujuan Cao},
  booktitle = {ICLR},
  year = 2026
}

@inproceedings{yu2025gamefactory,
  title={{GameFactory}: Creating new games with generative interactive videos},
  author={Yu, Jiwen and Qin, Yiran and Wang, Xintao and Wan, Pengfei and Zhang, Di and Liu, Xihui},
  booktitle = {ICCV},
  year = 2025
}

@inproceedings{he2024cameractrl,
  title={{CameraCtrl}: Enabling camera control for text-to-video generation},
  author={He, Hao and Xu, Yinghao and Guo, Yuwei and Wetzstein, Gordon and Dai, Bo and Li, Hongsheng and Yang, Ceyuan},
  booktitle = {ICLR},
  year = 2025
}

@inproceedings{wang2024motionctrl,
  title={{MotionCtrl}: A unified and flexible motion controller for video generation},
  author={Wang, Zhouxia and Yuan, Ziyang and Wang, Xintao and Li, Yaowei and Chen, Tianshui and Xia, Menghan and Luo, Ping and Shan, Ying},
  booktitle={ACM SIGGRAPH},
  pages={1--11},
  year={2024}
}

@inproceedings{miyato2023gta,
  title={{GTA}: A geometry-aware attention mechanism for multi-view transformers},
  author={Miyato, Takeru and Jaeger, Bernhard and Welling, Max and Geiger, Andreas},
  booktitle = {ICLR},
  year = 2024
}

@inproceedings{kong2024eschernet,
  title={{EscherNet}: A generative model for scalable view synthesis},
  author={Kong, Xin and Liu, Shikun and Lyu, Xiaoyang and Taher, Marwan and Qi, Xiaojuan and Davison, Andrew J},
  booktitle={CVPR},
  pages={9503--9513},
  year={2024}
}

@inproceedings{li2025cameras,
  title={Cameras as relative positional encoding},
  author={Li, Ruilong and Yi, Brent and Liu, Junchen and Gao, Hang and Ma, Yi and Kanazawa, Angjoo},
  booktitle={Advances in Neural Information Processing Systems},
  volume={38},
  pages={15984--16009},
  year={2025}
}

@inproceedings{bahmani2025ac3d,
  title={{AC3D}: Analyzing and improving {3D} camera control in video diffusion transformers},
  author={Bahmani, Sherwin and Skorokhodov, Ivan and Qian, Guocheng and Siarohin, Aliaksandr and Menapace, Willi and Tagliasacchi, Andrea and Lindell, David B and Tulyakov, Sergey},
  booktitle={CVPR},
  pages={22875--22889},
  year={2025}
}

@inproceedings{li2025vmem,
  title={{VMem}: Consistent Interactive Video Scene Generation with Surfel-Indexed View Memory},
  author={Li, Runjia and Torr, Philip and Vedaldi, Andrea and Jakab, Tomas},
  booktitle={ICCV},
  year={2025}
}

@inproceedings{yu2025wonderworld,
  title={{WonderWorld}: Interactive {3D} scene generation from a single image},
  author={Yu, Hong-Xing and Duan, Haoyi and Herrmann, Charles and Freeman, William T and Wu, Jiajun},
  booktitle={CVPR},
  pages={5916--5926},
  year={2025}
}

@inproceedings{ren2025gen3c,
  title={{Gen3C}: {3D}-informed world-consistent video generation with precise camera control},
  author={Ren, Xuanchi and Shen, Tianchang and Huang, Jiahui and Ling, Huan and Lu, Yifan and Nimier-David, Merlin and M{\"u}ller, Thomas and Keller, Alexander and Fidler, Sanja and Gao, Jun},
  booktitle={CVPR},
  pages={6121--6132},
  year={2025}
}

@inproceedings{cao2025uni3c,
  title={{Uni3C}: Unifying Precisely {3D}-Enhanced Camera and Human Motion Controls for Video Generation},
  author={Cao, Chenjie and Zhou, Jingkai and Li, Shikai and Liang, Jingyun and Yu, Chaohui and Wang, Fan and Xue, Xiangyang and Fu, Yanwei},
  booktitle={ACM SIGGRAPH Asia},
  year={2025}
}

@inproceedings{yu2025trajectorycrafter,
  title={{TrajectoryCrafter}: Redirecting camera trajectory for monocular videos via diffusion models},
  author={YU, Mark and Hu, Wenbo and Xing, Jinbo and Shan, Ying},
  booktitle={ICCV},
  year={2025}
}

@article{yu2024viewcrafter,
  title={{ViewCrafter}: Taming Video Diffusion Models for High-fidelity Novel View Synthesis},
  author={Yu, Wangbo and Xing, Jinbo and Yuan, Li and Hu, Wenbo and Li, Xiaoyu and Huang, Zhipeng and Gao, Xiangjun and Wong, Tien-Tsin and Shan, Ying and Tian, Yonghong},
  journal={IEEE Transactions on Pattern Analysis and Machine Intelligence},
  year={2025},
}

@inproceedings{xiao2025worldmem,
  title={{WorldMem}: Long-term consistent world simulation with memory},
  author={Xiao, Zeqi and Lan, Yushi and Zhou, Yifan and Ouyang, Wenqi and Yang, Shuai and Zeng, Yanhong and Pan, Xingang},
  booktitle={Advances in Neural Information Processing Systems},
  volume={38},
  pages={49632--49652},
  year={2026}
}

@inproceedings{yu2025context,
  title={Context as {Memory}: Scene-consistent interactive long video generation with memory retrieval},
  author={Yu, Jiwen and Bai, Jianhong and Qin, Yiran and Liu, Quande and Wang, Xintao and Wan, Pengfei and Zhang, Di and Liu, Xihui},
  booktitle={ACM SIGGRAPH Asia},
  year={2025}
}

@inproceedings{salimans2022progressive,
  title={Progressive distillation for fast sampling of diffusion models},
  author={Salimans, Tim and Ho, Jonathan},
  booktitle={ICLR},
  year={2022}
}

@inproceedings{frans2024one,
  title={One step diffusion via shortcut models},
  author={Frans, Kevin and Hafner, Danijar and Levine, Sergey and Abbeel, Pieter},
  booktitle={ICLR},
  year={2025}
}

@inproceedings{geng2025mean,
  title={Mean flows for one-step generative modeling},
  author={Geng, Zhengyang and Deng, Mingyang and Bai, Xingjian and Kolter, Zico and He, Kaiming},
  booktitle={Advances in Neural Information Processing Systems},
  volume={38},
  pages={75460--75482},
  year={2025}
}

@inproceedings{sauer2024fast,
  title={Fast high-resolution image synthesis with latent adversarial diffusion distillation},
  author={Sauer, Axel and Boesel, Frederic and Dockhorn, Tim and Blattmann, Andreas and Esser, Patrick and Rombach, Robin},
  booktitle={ACM SIGGRAPH Asia},
  pages={1--11},
  year={2024}
}

@article{lin2024sdxl,
  title={{SDXL-Lightning}: Progressive adversarial diffusion distillation},
  author={Lin, Shanchuan and Wang, Anran and Yang, Xiao},
  journal={arXiv preprint arXiv:2402.13929},
  year={2024}
}

@inproceedings{sauer2024adversarial,
  title={Adversarial diffusion distillation},
  author={Sauer, Axel and Lorenz, Dominik and Blattmann, Andreas and Rombach, Robin},
  booktitle={ECCV},
  pages={87--103},
  year={2024},
  organization={Springer}
}

@inproceedings{kang2024distilling,
  title={Distilling diffusion models into conditional gans},
  author={Kang, Minguk and Zhang, Richard and Barnes, Connelly and Paris, Sylvain and Kwak, Suha and Park, Jaesik and Shechtman, Eli and Zhu, Jun-Yan and Park, Taesung},
  booktitle={ECCV},
  pages={428--447},
  year={2024},
  organization={Springer}
}

@inproceedings{lin2025diffusion,
  title={Diffusion adversarial post-training for one-step video generation},
  author={Lin, Shanchuan and Xia, Xin and Ren, Yuxi and Yang, Ceyuan and Xiao, Xuefeng and Jiang, Lu},
  booktitle={ICML},
  year={2025},
}

@inproceedings{yin2024one,
  title={One-step diffusion with distribution matching distillation},
  author={Yin, Tianwei and Gharbi, Micha{\"e}l and Zhang, Richard and Shechtman, Eli and Durand, Fredo and Freeman, William T and Park, Taesung},
  booktitle={CVPR},
  pages={6613--6623},
  year={2024}
}

@inproceedings{yin2024improved,
  title={Improved distribution matching distillation for fast image synthesis},
  author={Yin, Tianwei and Gharbi, Micha{\"e}l and Park, Taesung and Zhang, Richard and Shechtman, Eli and Durand, Fredo and Freeman, Bill},
  booktitle={Advances in Neural Information Processing Systems},
  volume={37},
  pages={47455--47487},
  year={2024}
}

@inproceedings{lu2025adversarial,
  title={Adversarial distribution matching for diffusion distillation towards efficient image and video synthesis},
  author={Lu, Yanzuo and Ren, Yuxi and Xia, Xin and Lin, Shanchuan and Wang, Xing and Xiao, Xuefeng and Ma, Andy J and Xie, Xiaohua and Lai, Jian-Huang},
  booktitle={ICCV},
  pages={16818--16829},
  year={2025}
}

@inproceedings{wang2023prolificdreamer,
  title={{ProlificDreamer}: High-fidelity and diverse text-to-{3D} generation with variational score distillation},
  author={Wang, Zhengyi and Lu, Cheng and Wang, Yikai and Bao, Fan and Li, Chongxuan and Su, Hang and Zhu, Jun},
  booktitle={Advances in Neural Information Processing Systems},
  volume={36},
  pages={8406--8441},
  year={2023}
}

@inproceedings{yin2025slow,
  title={From slow bidirectional to fast autoregressive video diffusion models},
  author={Yin, Tianwei and Zhang, Qiang and Zhang, Richard and Freeman, William T and Durand, Fredo and Shechtman, Eli and Huang, Xun},
  booktitle={CVPR},
  pages={22963--22974},
  year={2025}
}

@inproceedings{huang2025self,
  title={{Self Forcing}: Bridging the train-test gap in autoregressive video diffusion},
  author={Huang, Xun and Li, Zhengqi and He, Guande and Zhou, Mingyuan and Shechtman, Eli},
  booktitle={Advances in Neural Information Processing Systems},
  volume={38},
  pages={167283--167308},
  year={2025}
}

@article{kingma2013auto,
  title={Auto-encoding variational bayes},
  author={Kingma, Diederik P and Welling, Max},
  journal={arXiv preprint arXiv:1312.6114},
  year={2013}
}

@article{su2024roformer,
  title={{RoFormer}: Enhanced transformer with rotary position embedding},
  author={Su, Jianlin and Ahmed, Murtadha and Lu, Yu and Pan, Shengfeng and Bo, Wen and Liu, Yunfeng},
  journal={Neurocomputing},
  volume={568},
  year={2024},
  publisher={Elsevier}
}

@inproceedings{zhou2025stable,
  title={{Stable Virtual Camera}: Generative view synthesis with diffusion models},
  author={Zhou, Jensen and Gao, Hang and Voleti, Vikram and Vasishta, Aaryaman and Yao, Chun-Han and Boss, Mark and Torr, Philip and Rupprecht, Christian and Jampani, Varun},
  booktitle={ICCV},
  pages={12405--12414},
  year={2025}
}

@inproceedings{li2025sekai,
  title={Sekai: A video dataset towards world exploration},
  author={Li, Zhen and Li, Chuanhao and Mao, Xiaofeng and Lin, Shaoheng and Li, Ming and Zhao, Shitian and Xu, Zhaopan and Li, Xinyue and Feng, Yukang and Sun, Jianwen and others},
  booktitle={Advances in Neural Information Processing Systems},
  volume={38},
  year={2025}
}

@inproceedings{ling2024dl3dv,
  title={{DL3DV-10K}: A large-scale scene dataset for deep learning-based {3D} vision},
  author={Ling, Lu and Sheng, Yichen and Tu, Zhi and Zhao, Wentian and Xin, Cheng and Wan, Kun and Yu, Lantao and Guo, Qianyu and Yu, Zixun and Lu, Yawen and others},
  booktitle={CVPR},
  pages={22160--22169},
  year={2024}
}

@inproceedings{wu2025difix3d+,
  title={{Difix3D+}: Improving {3D} reconstructions with single-step diffusion models},
  author={Wu, Jay Zhangjie and Zhang, Yuxuan and Turki, Haithem and Ren, Xuanchi and Gao, Jun and Shou, Mike Zheng and Fidler, Sanja and Gojcic, Zan and Ling, Huan},
  booktitle={CVPR},
  pages={26024--26035},
  year={2025}
}

@inproceedings{yang2025longlive,
  title={{LongLive}: Real-time interactive long video generation},
  author={Yang, Shuai and Huang, Wei and Chu, Ruihang and Xiao, Yicheng and Zhao, Yuyang and Wang, Xianbang and Li, Muyang and Xie, Enze and Chen, Yingcong and Lu, Yao and others},
  booktitle={ICLR},
  year={2026}
}

@article{hunyuanworld2025tencent,
    title={{HunyuanWorld 1.0}: Generating Immersive, Explorable, and Interactive {3D} Worlds from Words or Pixels},
    author={Team HunyuanWorld},
    journal={arXiv preprint arXiv:2507.21809},
    year={2025}
}

@article{liu2025worldmirror,
  title={{WorldMirror}: Universal {3D} World Reconstruction with Any-Prior Prompting},
  author={Liu, Yifan and Min, Zhiyuan and Wang, Zhenwei and Wu, Junta and Wang, Tengfei and Yuan, Yixuan and Luo, Yawei and Guo, Chunchao},
  journal={arXiv preprint arXiv:2510.10726},
  year={2025}
}

@inproceedings{zhang2025sageattention,
  title={{SageAttention}: Accurate 8-Bit Attention for Plug-and-play Inference Acceleration}, 
  author={Zhang, Jintao and Wei, Jia and Zhang, Pengle and Zhu, Jun and Chen, Jianfei},
  booktitle={ICLR},
  year={2025}
}

@inproceedings{che2024gamegen,
  title={{GameGen-X}: Interactive open-world game video generation},
  author={Che, Haoxuan and He, Xuanhua and Liu, Quande and Jin, Cheng and Chen, Hao},
  booktitle={ICLR},
  year={2025}
}

@inproceedings{sun2024dimensionx,
  title={{DimensionX}: Create any {3D} and {4D} scenes from a single image with controllable video diffusion},
  author={Sun, Wenqiang and Chen, Shuo and Liu, Fangfu and Chen, Zilong and Duan, Yueqi and Zhang, Jun and Wang, Yikai},
  booktitle={ICCV},
  year={2025}
}

@inproceedings{li-etal-2023-sequence,
    title = "Sequence Parallelism: Long Sequence Training from System Perspective",
    author = "Li, Shenggui  and
      Xue, Fuzhao  and
      Baranwal, Chaitanya  and
      Li, Yongbin  and
      You, Yang",
    editor = "Rogers, Anna  and
      Boyd-Graber, Jordan  and
      Okazaki, Naoaki",
    booktitle = "Proceedings of the 61st Annual Meeting of the Association for Computational Linguistics (Volume 1: Long Papers)",
    month = jul,
    year = "2023",
    address = "Toronto, Canada",
    publisher = "Association for Computational Linguistics",
    url = "https://aclanthology.org/2023.acl-long.134/",
    doi = "10.18653/v1/2023.acl-long.134",
    pages = "2391--2404",
}

@article{mao2025yume,
  title={{YUME}: An interactive world generation model},
  author={Mao, Xiaofeng and Lin, Shaoheng and Li, Zhen and Li, Chuanhao and Peng, Wenshuo and He, Tong and Pang, Jiangmiao and Chi, Mingmin and Qiao, Yu and Zhang, Kaipeng},
  journal={arXiv preprint arXiv:2507.17744},
  year={2025}
}

@article{zhang2025accvideo,
  title={{AccVideo}: Accelerating video diffusion model with synthetic dataset},
  author={Zhang, Haiyu and Chen, Xinyuan and Wang, Yaohui and Liu, Xihui and Wang, Yunhong and Qiao, Yu},
  journal={arXiv preprint arXiv:2503.19462},
  year={2025}
}

@inproceedings{lin2025autoregressive,
  title={Autoregressive adversarial post-training for real-time interactive video generation},
  author={Lin, Shanchuan and Yang, Ceyuan and He, Hao and Jiang, Jianwen and Ren, Yuxi and Xia, Xin and Zhao, Yang and Xiao, Xuefeng and Jiang, Lu},
  booktitle={Advances in Neural Information Processing Systems},
  volume={38},
  pages={41061--41086},
  year={2025}
}

@article{huang2025vipe,
  title={{VIPE}: Video pose engine for {3D} geometric perception},
  author={Huang, Jiahui and Zhou, Qunjie and Rabeti, Hesam and Korovko, Aleksandr and Ling, Huan and Ren, Xuanchi and Shen, Tianchang and Gao, Jun and Slepichev, Dmitry and Lin, Chen-Hsuan and others},
  journal={arXiv preprint arXiv:2508.10934},
  year={2025}
}

@article{liu2024reconx,
  title={{ReconX}: Reconstruct any scene from sparse views with video diffusion model},
  author={Liu, Fangfu and Sun, Wenqiang and Wang, Hanyang and Wang, Yikai and Sun, Haowen and Ye, Junliang and Zhang, Jun and Duan, Yueqi},
  journal={IEEE Transactions on Image Processing},
  year={2026},
  publisher={IEEE}
}

@article{sun2025virtual,
  title={From Virtual Games to Real-World Play},
  author={Sun, Wenqiang and Wei, Fangyun and Zhao, Jinjing and Chen, Xi and Chen, Zilong and Zhang, Hongyang and Zhang, Jun and Lu, Yan},
  journal={arXiv preprint arXiv:2506.18901},
  year={2025}
}

@inproceedings{liu2025rolling,
  title={{Rolling Forcing}: Autoregressive Long Video Diffusion in Real Time},
  author={Liu, Kunhao and Hu, Wenbo and Xu, Jiale and Shan, Ying and Lu, Shijian},
  booktitle={ICLR},
  year={2026}
}

@inproceedings{cui2025self,
  title={{Self-Forcing++}: Towards Minute-Scale High-Quality Video Generation},
  author={Cui, Justin and Wu, Jie and Li, Ming and Yang, Tao and Li, Xiaojie and Wang, Rui and Bai, Andrew and Ban, Yuanhao and Hsieh, Cho-Jui},
  booktitle={ICLR},
  year={2026}
}

@inproceedings{Redmon_2016_CVPR,
  title={You only look once: Unified, real-time object detection},
  author={Redmon, Joseph and Divvala, Santosh and Girshick, Ross and Farhadi, Ali},
  booktitle={CVPR},
  pages={779--788},
  year={2016}
}

@inproceedings{huang2024vbench,
  title={{VBench}: Comprehensive benchmark suite for video generative models},
  author={Huang, Ziqi and He, Yinan and Yu, Jiashuo and Zhang, Fan and Si, Chenyang and Jiang, Yuming and Zhang, Yuanhan and Wu, Tianxing and Jin, Qingyang and Chanpaisit, Nattapol and others},
  booktitle={CVPR},
  pages={21807--21818},
  year={2024}
}

@inproceedings{duan2025worldscore,
  title={{WorldScore}: A unified evaluation benchmark for world generation},
  author={Duan, Haoyi and Yu, Hong-Xing and Chen, Sirui and Fei-Fei, Li and Wu, Jiajun},
  booktitle={ICCV},
  pages={27713--27724},
  year={2025}
}

@inproceedings{xiao2023efficient,
    title={Efficient streaming language models with attention sinks},
    author={Xiao, Guangxuan and Tian, Yuandong and Chen, Beidi and Han, Song and Lewis, Mike},
    booktitle = {ICLR},
    year = 2024
}

@inproceedings{chen2025learning,
  title={Learning world models for interactive video generation},
  author={Chen, Taiye and Hu, Xun and Ding, Zihan and Jin, Chi},
  booktitle={Advances in Neural Information Processing Systems},
  volume={38},
  pages={154456--154483},
  year={2025}
}

@article{mao2025yume2,
  title={{YUME-1.5}: A Text-Controlled Interactive World Generation Model},
  author={Mao, Xiaofeng and Li, Zhen and Li, Chuanhao and Xu, Xiaojie and Ying, Kaining and He, Tong and Pang, Jiangmiao and Qiao, Yu and Zhang, Kaipeng},
  journal={arXiv preprint arXiv:2512.22096},
  year={2025}
}

@article{xiang2025pan,
  title={{PAN}: A World Model for General, Interactable, and Long-Horizon World Simulation},
  author={Xiang, Jiannan and Gu, Yi and Liu, Zihan and Feng, Zeyu and Gao, Qiyue and Hu, Yiyan and Huang, Benhao and Liu, Guangyi and Yang, Yichi and Zhou, Kun and others},
  journal={arXiv preprint arXiv:2511.09057},
  year={2025}
}

@article{tang2025hunyuan,
  title={{Hunyuan-GameCraft-2}: Instruction-following Interactive Game World Model},
  author={Tang, Junshu and Liu, Jiacheng and Li, Jiaqi and Wu, Longhuang and Yang, Haoyu and Zhao, Penghao and Gong, Siruis and Yuan, Xiang and Shao, Shuai and Lu, Qinglin},
  journal={arXiv preprint arXiv:2511.23429},
  year={2025}
}

@inproceedings{yesiltepe2025infinity,
  title={{Infinity-RoPE}: Action-Controllable Infinite Video Generation Emerges From Autoregressive Self-Rollout},
  author={Yesiltepe, Hidir and Meral, Tuna Han Salih and Akan, Adil Kaan and Oktay, Kaan and Yanardag, Pinar},
  booktitle={CVPR},
  year={2026}
}

@article{hong2025relic,
  title={{RELIC}: Interactive Video World Model with Long-Horizon Memory},
  author={Hong, Yicong and Mei, Yiqun and Ge, Chongjian and Xu, Yiran and Zhou, Yang and Bi, Sai and Hold-Geoffroy, Yannick and Roberts, Mike and Fisher, Matthew and Shechtman, Eli and others},
  journal={arXiv preprint arXiv:2512.04040},
  year={2025}
}

@inproceedings{zhang2025packing,
  title={Frame context packing and drift prevention in next-frame-prediction video diffusion models},
  author={Zhang, Lvmin and Cai, Shengqu and Li, Muyang and Wetzstein, Gordon and Agrawala, Maneesh},
  booktitle={Advances in Neural Information Processing Systems},
  volume={38},
  pages={30546--30566},
  year={2025}
}

@inproceedings{yu2023wonderjourney,
  title={{WonderJourney}: Going from anywhere to everywhere},
  author={Yu, Hong-Xing and Duan, Haoyi and Hur, Junhwa and Sargent, Kyle and Rubinstein, Michael and Freeman, William T and Cole, Forrester and Sun, Deqing and Snavely, Noah and Wu, Jiajun and others},
  booktitle={CVPR},
  pages={6658--6667},
  year={2024}
}

@inproceedings{xu2024easyanimate,
  title={{EasyAnimate}: A high-performance long video generation method based on transformer architecture},
  author={Xu, Jiaqi and Zou, Xinyi and Huang, Kunzhe and Chen, Yunkuo and Liu, Bo and Cheng, MengLi and Shi, Xing and Huang, Jun},
  booktitle={ACM MM},
  year={2025}
}

@article{allegro2024,
  title={Allegro: Open the Black Box of Commercial-Level Video Generation Model},
  author={Yuan Zhou and Qiuyue Wang and Yuxuan Cai and Huan Yang},
  journal={arXiv preprint arXiv:2410.15458},
  year={2024}
}

@article{gen3,
  title={Introducing {GEN-3} alpha: A new frontier for video gneration},
  author={Runway},
  journal={},
  url={https://runwayml.com/research/introducing-gen-3-alpha},
  year={2024}
}

@article{huang2025voyager,
  title={Voyager: Long-range and world-consistent video diffusion for explorable {3D} scene generation},
  author={Huang, Tianyu and Zheng, Wangguandong and Wang, Tengfei and Liu, Yuhao and Wang, Zhenwei and Wu, Junta and Jiang, Jie and Li, Hui and Lau, Rynson and Zuo, Wangmeng and others},
  journal={ACM TOG},
  volume={44},
  number={6},
  pages={1--15},
  year={2025},
  publisher={ACM New York, NY, USA}
}

@article{cui2026lol,
  title={{LoL}: Longer than Longer, Scaling Video Generation to Hour},
  author={Cui, Justin and Wu, Jie and Li, Ming and Yang, Tao and Li, Xiaojie and Wang, Rui and Bai, Andrew and Ban, Yuanhao and Hsieh, Cho-Jui},
  journal={arXiv preprint arXiv:2601.16914},
  year={2026}
}

@inproceedings{asim2025met3r,
  title={{MEt3R}: Measuring multi-view consistency in generated images},
  author={Asim, Mohammad and Wewer, Christopher and Wimmer, Thomas and Schiele, Bernt and Lenssen, Jan Eric},
  booktitle={CVPR},
  pages={6034--6044},
  year={2025}
}

@InProceedings{Wang_2024_CVPR,
    author    = {Wang, Shuzhe and Leroy, Vincent and Cabon, Yohann and Chidlovskii, Boris and Revaud, Jerome},
    title     = {{DUSt3R}: Geometric {3D} Vision Made Easy},
    booktitle = {CVPR},
    year      = {2024},
    pages     = {20697-20709}
}

@inproceedings{shin2025motionstream,
  title={{MotionStream}: Real-time video generation with interactive motion controls},
  author={Shin, Joonghyuk and Li, Zhengqi and Zhang, Richard and Zhu, Jun-Yan and Park, Jaesik and Shechtman, Eli and Huang, Xun},
  booktitle={ICLR},
  year={2026}
}
\bibliographystyle{icml2026}


\newpage
\appendix
\onecolumn

\section{Training and Inference Details}
\label{supp_A}
We adopt the pretrained DiT-based video diffusion models~\cite{wan2025wan,kong2024hunyuanvideo} as the backbone. For the chunk-wise autoregressive diffusion transformer, we group 4 latents into a chunk. For the memory context, we set the temporal memory length to 3 chunks and the spatial memory length to 1 chunk. Moreover, inspired by~\citep{yang2025longlive, liu2025rolling, xiao2023efficient}, we observe that preserving the first chunk as attention sink enhances long-term consistency and further mitigates error accumulation. For the bidirectional teacher model $V_{\beta}$, we also adopt the dual-action representation and construct the memory context as described in Sec.~\ref{sec:distill}. The training consists of three stages.

\noindent \textbf{Stage One: Action Control.} In the first stage, we focus on injecting action control into the pretrained model. We employ the dual action representation to the pretrained model and train the bidirectional action model for 30K iterations. Then, we replace the 3D self-attention with block causal attention and train for an additional 30K iterations as our AR action model. We find that this enables the AR action model to converge more easily. In this stage, the model is trained on 61 frames (4 chunks) using the Adam optimizer with a learning rate of $1e-5$ and a batch size of 64.

\noindent \textbf{Stage Two: Memory.} In the second stage, we train the bidirectional action model and the AR action model with context memory as described in Sec.3.3 and Sec.3.4, respectively. \wenq{For the bidirectional action model, the generation sequence consists of 4 chunks $x_{j:j+3}$ (16 latents, 61 frames). We utilize $C_{j:j+3} - x_{j:j+3}$ as a variable-length memory context for the teacher. Following the Flow Matching framework, we apply noise from $[0, 1]$ to $x_{j:j+3}$, while the memory context $C_{j:j+3} - x_{j:j+3}$ undergoes uniform noise sampling from $[0, 0.2]$ to enable robust conditioning. Crucially, the training loss are computed strictly on the generation sequence. In stage two, both the bidirectional and AR models are trained on sequences of up to 160 latents (637 frames).} Other settings remain the same as in the first stage.

\noindent \textbf{Stage Three: Context Forcing.} In the final stage, we use the bidirectional model as the teacher and the AR model as the student for distillation. To stabilize the distillation process, we employ a progressive training strategy that gradually increases the maximum length of the generated latents. For the student model, the learning rate is set to $1e-6$, while for the bidirectional model, which is used to compute the fake score, the learning rate is set to $2e-7$. The models are trained for 2K iteration with a batch size of 64. All other hyperparameters follow~\cite{huang2025self}. For the details of context forcing, see Algorithm~\ref{alg:context_forcing}.

Finally, our AR model can produce multiple chunks in a streaming fashion with KV cache as shown in Algorithm~\ref{alg:inference}. When the user provides only camera poses, we first compute the relative translations and rotations between consecutive poses, and then apply a thresholding mechanism to identify and convert them into discrete actions. Conversely, when only discrete actions are available, we use the predefined relative translations and rotations associated with each action to convert them into camera poses. 

\begin{figure}[htp]
\vspace{-2em}
\begin{minipage}[t]{0.48\textwidth}
  \begin{algorithm}[H]
    \caption{Context Forcing Training}
    \small
    \begin{algorithmic}[1]
      \Require Number of denoising timesteps $d$ and chunks $n=4$
      \Require Dataset $D$ (encoded by 3D VAE)
      \Require AR diffusion model $N_\theta$
      \Require Bidirectional diffusion model $V_\beta^{fake}$ and $V^{real}$
      \Loop
        \State Progressively increase maximum chunk length $m$
        \State Sample chunk length $j \sim \text{Uniform}(0,1,\ldots, m)$
        \State Sample context $x_{0:j-1} \sim D$ 
        \For{$i = j, \dots, j+n-1$}
            \State Initialize $x_{i}^{init} \sim \mathcal{N}(0, I)$
            \State Reconstitute context memory $C_{i} \subseteq \{x_{0},\ldots,x_{i-1}\}$ 
            \State Sample $s \sim \text{Uniform}(1, 2, \ldots, d)$
            
            \State Self-rollout $x_{i}$ using $N_{\theta}$ with $C_{i}$ and $s$ denoising steps
        \EndFor
        \State Align context memory $C^{tea} \gets C_{j:j+n-1}-x_{j:j+n-1}$
        \State Sample diffusion timestep $k \sim [0,1]$
        \State $\hat{x}_{j:j+n-1} \gets \textit{AddNoise}({x}_{j:j+n-1},k)$
        \State Fake score $S^{fake} \gets V_{\beta}^{fake}(\hat{x}_{j:j+n-1},C^{tea}, k) $
        \State Real score $S^{fake} \gets V^{real}(\hat{x}_{j:j+n-1},C^{tea}, k)$
        \State Update $\theta$ via distribution matching loss
    
        \State Update $\beta$ via flow matching loss~\cite{huang2025self}
    \EndLoop
    \end{algorithmic}
    \label{alg:context_forcing}
  \end{algorithm}
\end{minipage}
\hfill
\begin{minipage}[t]{0.48\textwidth}
  \begin{algorithm}[H]
    \caption{Inference with KV Cache}
    \small
    \begin{algorithmic}[1]
      \Require Number of inference chunks $n_{c}$
      \Require Denoise timesteps $\{k_1, \dots, k_d\}$
      \Require Number of inference chunks $n_{c}$
      \Require AR diffusion model $N_\theta$ (returns KV embeddings via $N_{\theta}^{\text{KV}}$)
      \State Initialize model output $X_{\theta} \gets []$
      \State Initialize KV cache $\KVSet \gets []$
      \For{$i = 0, \dots, n_{c} -1$}
         \State Initialize $x_{i} \sim \mathcal{N}(0, I)$
         \State Reconstitute context memory $C_{i} \subseteq \{x_{0},\ldots,x_{i-1}\}$ 
         \For{$s = d, \dots, 1$}
            \If{$s = d$ and $i > 1$}
                \State Reset $\KVSet \gets N_{\theta}^{\text{KV}}(C_{i}, 0)$
            \EndIf
            \State Denoise $x_{i} \gets N_{\theta}(x_{i}, \KVSet, k_{s})$
         \EndFor
         \State Add output $X_{\theta}$.append($x_{i}$)
      \EndFor
      \State \Return $X_{\theta}$
    \end{algorithmic}
    \label{alg:inference}
  \end{algorithm}
\end{minipage}
\vspace{-1em}
\end{figure}

\begin{table*}[ht!]
\caption{\textbf{Data organization.} The table details the four categories of data, their sources, the availability of action annotations (discrete and continuous), the number of clips, and their corresponding ratio in the final dataset.}
\label{tab:dataset_component}
\centering
    \newcommand{\gou}{\textcolor{ForestGreen}{\ding{52}}}
    \newcommand{\cha}{\textcolor{Red}{\ding{55}}}
\scalebox{0.8}{
\begin{tabular}{p{0.25\textwidth} >{\centering}p{0.3\textwidth}> {\centering}p{0.3\textwidth} >{\centering}p{0.1\textwidth} > {\centering\arraybackslash}p{0.1\textwidth}}
\toprule
\textbf{Category} & \textbf{Data Source} & \textbf{Annotation (discrete, continuous)} & \textbf{\# Clips} & \textbf{Ratios} \\
\midrule
{Real-World Dynamics} & Sekai \cite{li2025sekai} & (\cha, \cha) & 40K & 12.5\% \\
\midrule
Real-World 3D Scene & DL3DV \cite{ling2024dl3dv} & (\gou, \gou) & 60K & 18.75\% \\
\midrule
Synthetic 3D Scene & UE Rendering & (\cha, \gou) & 50K & 15.625\% \\
\midrule
Simulation Dynamics & Game Video Recordings & (\gou, \cha) & 170K & 53.125\% \\
\bottomrule
\end{tabular}
}
\vskip -0.12in
\end{table*}

\begin{figure*}
  \centering
  \includegraphics[width=1\linewidth]{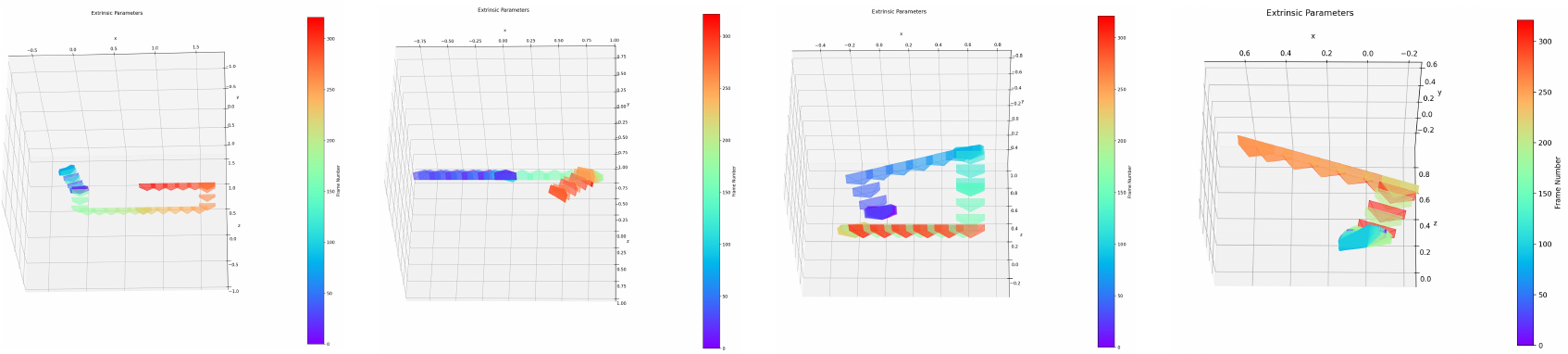}

  \caption{\textbf{Camera trajectories included in our collected dataset.} 
  }
  \label{fig:pose_vis}
\end{figure*}
  
\section{Dataset}
\label{supp_B}
Table~\ref{tab:dataset_component} provides a comprehensive breakdown of our dataset. We deliberately curate a diverse and high-quality collection, encompassing data from the simulation engine and real world, as well as static and dynamic environments, to guarantee the strong generalization of our model.

For Real-World Dynamics, we employ the Sekai dataset~\cite{li2025sekai}. However, the original videos often suffer from scene clutter and high dynamics. 
To address these issues, we implement a rigorous filtering pipeline. Specifically, we apply a state-of-the-art object detection model (YOLO \cite{Redmon_2016_CVPR}) to identify the presence of crowds and vehicles. By setting an empirical threshold, we filter out clips with high densities of moving objects, thereby ensuring annotation accuracy and stable training.

Regarding the Real-World 3D Scene data (DL3DV~\cite{ling2024dl3dv}), the original videos lack diversity in camera movement speed and trajectory complexity. To overcome this, we implement a sophisticated processing workflow: 3D Scene Reconstruction $\rightarrow$ Customized Trajectory Rendering $\rightarrow$ Visual Quality Filtering $\rightarrow$ Video Repair Post-processing (using Difix3D+~\cite{wu2025difix3d+}). This procedure yields additional 60K high-quality real video clips featuring balanced movement speed. During the customized trajectory rendering stage, we deliberately design diverse revisit trajectories to facilitate the learning of long-term geometric consistency. The discrete actions and continuous camera poses in these rendered data are highly accurate, which helps the model learn well-structured action patterns. 

For Synthetic 3D Scene (UE Rendering) data, we collect hundreds of UE scenes and obtain 50K video clips by rendering complex, customized trajectories. For Simulation Dynamics (Game Video Recordings), we establish a dedicated game recording platform and invite players to record 170K video clips from 1st/3rd-person AAA games.

We segment the original long videos into 30 to 40 seconds clips and employ a vision-language model to produce descriptive text annotations for every clip. Subsequently, we leverage VIPE~\cite{huang2025vipe} to generate high-quality camera poses for clips without camera annotations. However, given the long duration and high scene diversity of our dataset, we observe that pose estimation could be inaccurate, \ie, pose collapse. Therefore, we filter out videos whose adjacent frames exhibit erratic camera positions or rotation angles. \wenq{Specifically, we utilize the Peak-to-Median Ratio (PMR, the ratio of the maximum inter-frame motion to the median) of inter-frame motions as a detection metric. We subsample the predicted poses every four frames and compute the relative transformations to approximate the camera's instantaneous velocity. By evaluating the PMR, we can robustly identify impulsive pose jumps. Clips exhibiting a PMR above a conservative threshold (set to 5.0 in our experiments) are classified as unacceptable and are discarded.} Finally, for clips lacking discrete action annotations, we derive them from the continuous camera poses: we project the rotation and translation components onto the $x, y, z$ axes and apply a threshold to map these continuous values into corresponding discrete action states.

Fig.~\ref{fig:pose_vis} illustrates the camera trajectories. Our dataset contains complex and diverse trajectories, including a large number of revisit trajectories, which enables our model to learn precise action control and long-term geometric consistency.

\section{Additional Experimental Results}

\begin{figure*}
  \centering
  \includegraphics[width=1\linewidth]{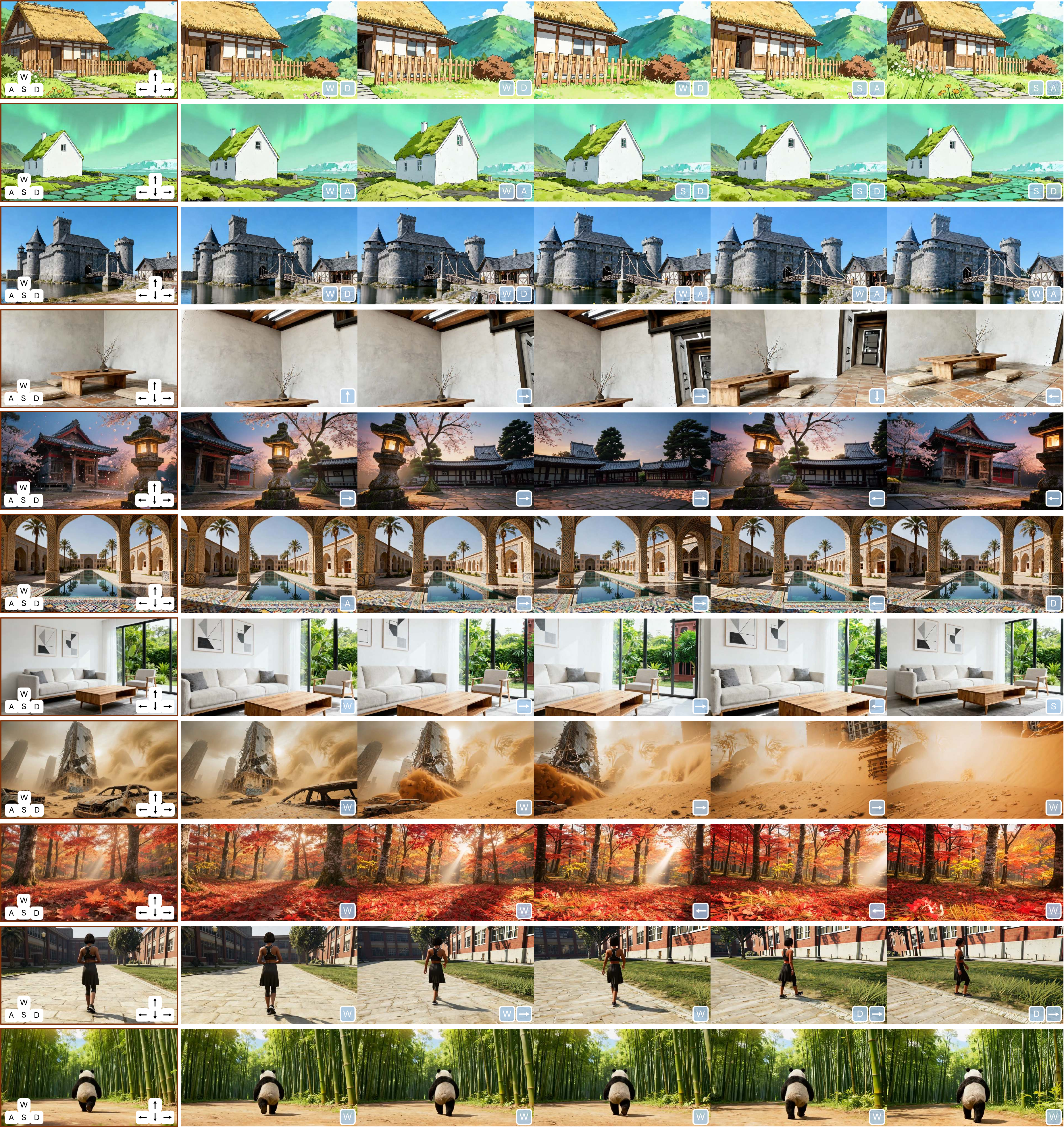}

  \caption{\textbf{More qualitative results.}
  }
  \label{fig:complex_action}
\end{figure*}

\begin{figure*}
  \centering
  \includegraphics[width=1\linewidth]{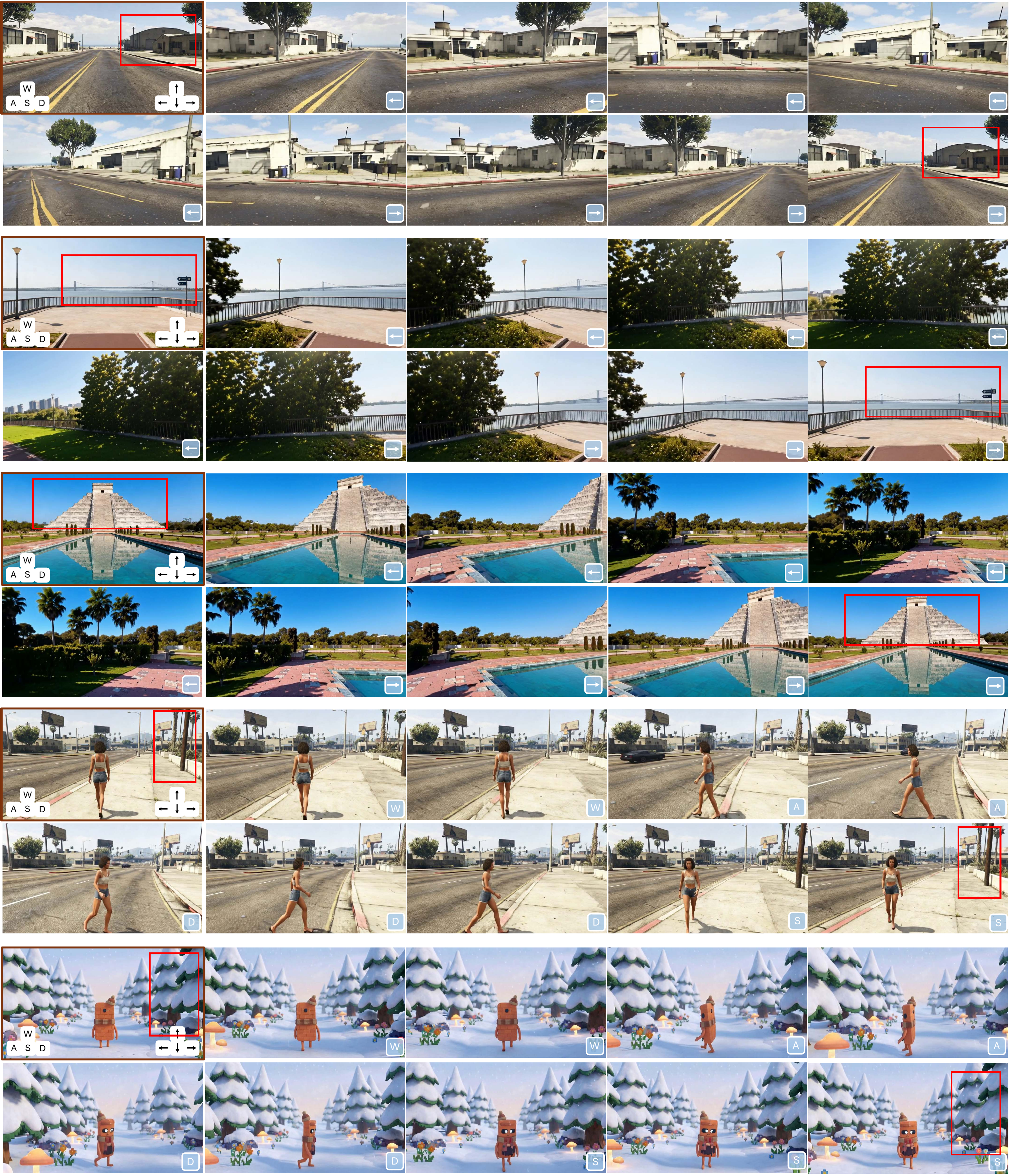}

  \caption{\textbf{More visualization results.}
  }
  \label{fig:large_motion}
\end{figure*}

\begin{figure*}
  \centering
  \includegraphics[width=0.9\linewidth]{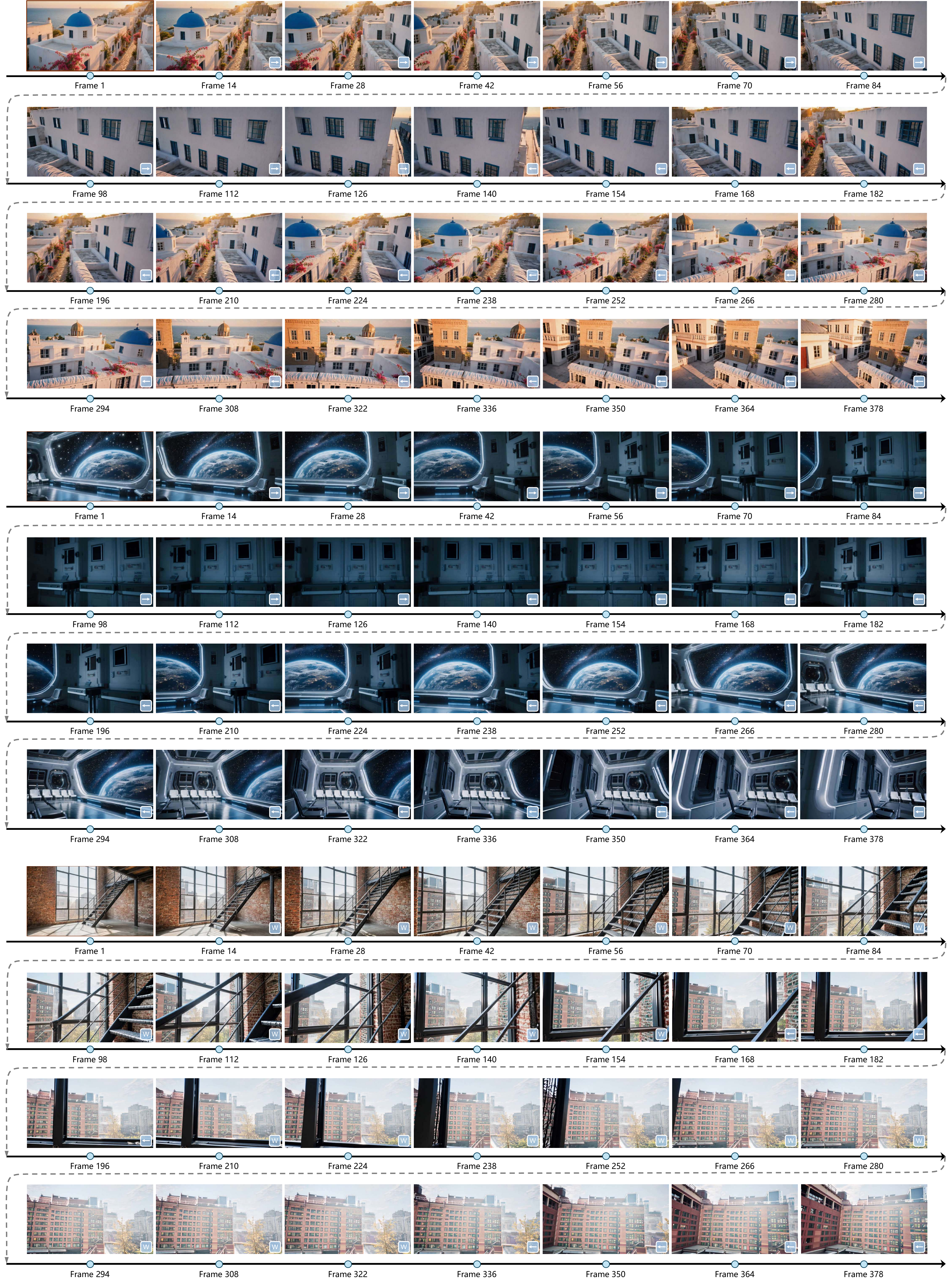}

  \caption{\textbf{Long video generation.} 
  }
  \label{fig:long_video}
\end{figure*}

\subsection{More Qualitative Results}
Fig.~\ref{fig:complex_action} illustrates the results of WorldPlay under various actions and virtual environments. As shown in the first three rows, we can interact with complex composite actions, \eg, various combinations of movements. Moreover, WorldPlay can follow intricate trajectories, such as complex rotations and alternating sequences of rotations and movements as demonstrated in the middle six rows. This enhanced control capability is enabled by our dual action representation, which allows for more precise and reliable action guidance.
Furthermore, WorldPlay exhibits strong generalization, enabling it to control different types of agents, \eg, human or animals, to roam within the scenes as shown \wenq{in the last two rows in Fig.~\ref{fig:complex_action} and the last two cases in Fig.~\ref{fig:large_motion}}. For more intuitive perspectives, please refer to the supplementary videos. 

\subsection{Long Video Generation}
\begin{wrapfigure}{r}{0.5\textwidth}
  \centering
  \vspace{-8mm}
  \includegraphics[width=0.48\textwidth]{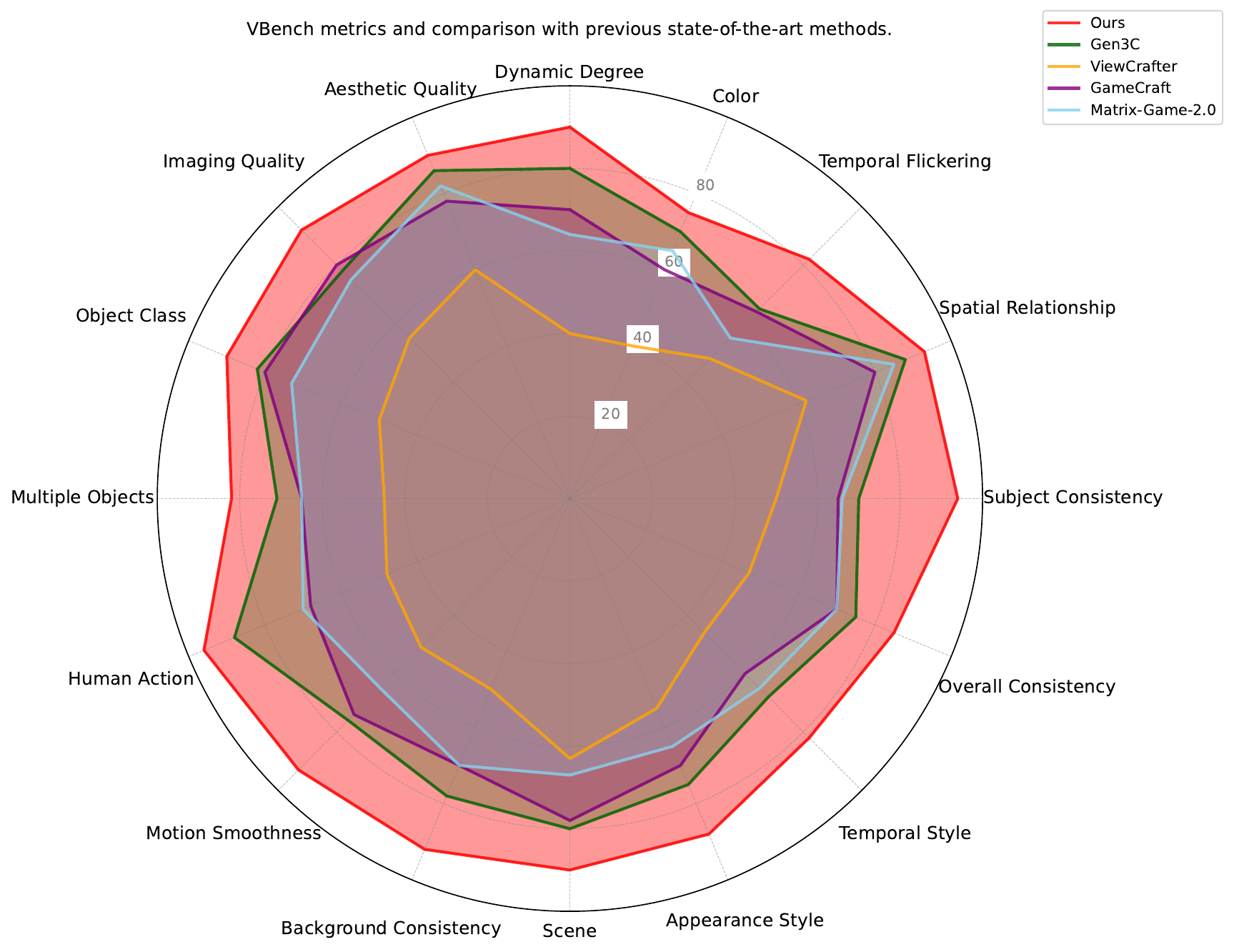}
  \caption{\textbf{VBench evaluation.}}
  \label{fig:vbench_plot}
  \vspace{-10mm}
\end{wrapfigure}
Fig.~\ref{fig:long_video} presents long video generation results from WorldPlay, we maintain long-term consistency, \eg, frame 1 and frame 252 in the top two examples, and preserve high visual quality throughout the entire sequence. Moreover, our context memory ensures that the generation time for each chunk remains constant and does not increase as the video length grows, enabling real-time interactivity and enhancing the user’s immersive experience. \wenq{Furthermore, the first three rows of Fig.~\ref{fig:large_motion} illustrate the generated results spanning 637 frames.}

\subsection{Comparison of Models under Context Forcing}

\begin{figure}
  \centering
  \includegraphics[width=1\linewidth]{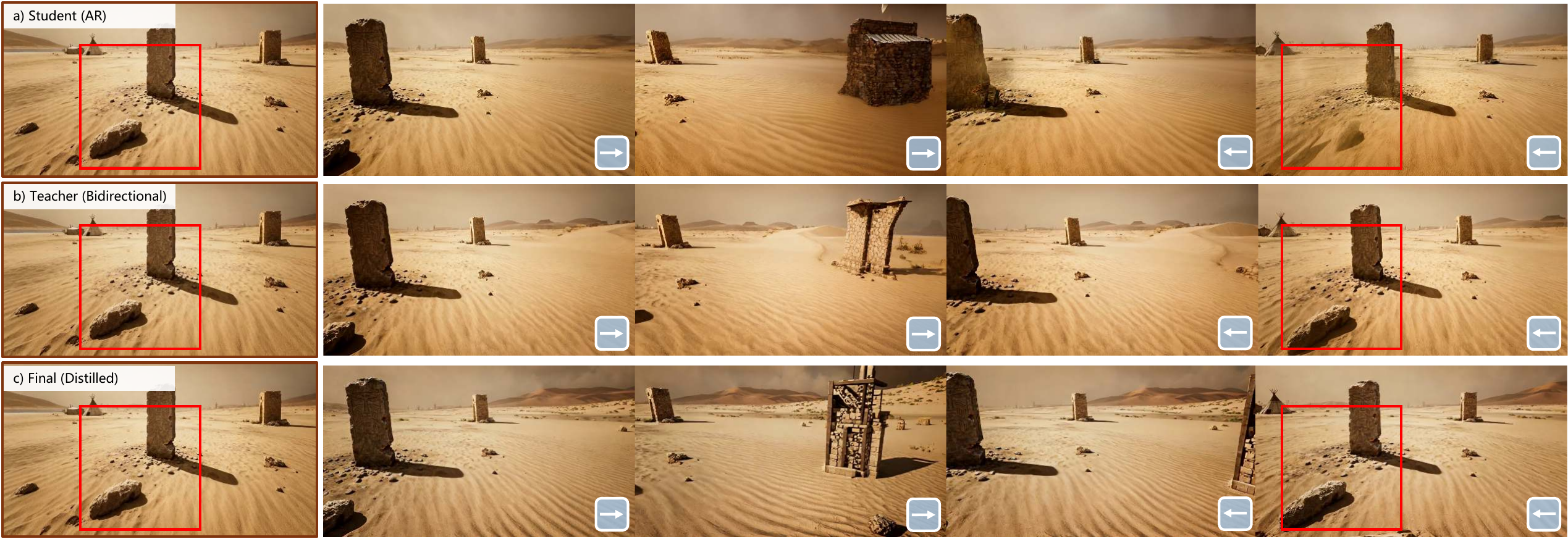}
  \caption{Visualization of different models under context forcing.
  }
  \label{fig:context_forcing_comp}
  \vspace{-2em}
\end{figure}

We provide a comprehensive comparison of different models under context forcing in Table~\ref{tab:ablation_distill_context} and Fig.~\ref{fig:context_forcing_comp}. The teacher model exhibits better control capability and visual quality due to the bidirectional nature, which provides reliable guidance during distillation. However, this limits its real-time interactivity. Through context forcing, we mitigate error accumulation while maintaining and even surpassing long-term consistency of the student model, yielding improved overall performance. In addition, context forcing reduces the student model’s inference steps, enabling real-time interaction.

\begin{table}[t]
    \caption{
    \textbf{Left: comparison of Models under Context Forcing.} The results are evaluated on the long-term test data. Student (AR) denotes the AR model before distillation, Teacher (bidirectional) refers to the memory-augmented bidirectional video diffusion model, and Final (distilled) represents the AR model after distillation. NFE denotes the number of function evaluations.  \textbf{Right: ablation for memory size.} Spa. and Tem. denote the number of chunks in spatial memory and temporal memory, respectively.
    }
    \label{tab:ablation_distill_context}
  \vspace{-2mm}
  \centering
  \begin{minipage}[t]{0.485\textwidth}
    \centering
    \footnotesize
    \setlength{\tabcolsep}{2pt}
    \resizebox{1.05\textwidth}{!}{
    \begin{tabular}{l|c|ccccc}
        \toprule
        \textbf{} & \textbf{NFE} & \textbf{PSNR$\uparrow$} & \textbf{SSIM$\uparrow$} & \textbf{LPIPS$\downarrow$} & $R_{\text{dist}}$ $\downarrow$ & $T_{\text{dist}}$ $\downarrow$ \\
        \midrule
        Student (AR)  & 100 & {16.27} & {0.425} & {0.495} & {0.611} & {0.991} \\
        Teacher (Bidirectional)  & 100 & \textbf{19.31} & \textbf{0.599} & 0.383 & \textbf{0.209} & \textbf{0.717}  \\
        Final (Distilled) & 4 & {18.94} & {0.585} &  \textbf{0.371}  & {0.332} & {0.797}  \\
        \bottomrule
    \end{tabular}
    }
  \end{minipage}
  \hfill
  \begin{minipage}[t]{0.485\textwidth}
    \centering
    \footnotesize
    \setlength{\tabcolsep}{2pt}
    \scriptsize 
    \resizebox{0.95\textwidth}{!}{
    \begin{tabular}{cc|ccccc}
        \toprule
        \textbf{Spa.} & \textbf{Tem.} & \textbf{PSNR$\uparrow$} & \textbf{SSIM$\uparrow$} & \textbf{LPIPS$\downarrow$} & $R_{\text{dist}}$ $\downarrow$ & $T_{\text{dist}}$ $\downarrow$ \\
        \midrule
        3 & 1  & \textbf{16.41} & 0.418 & 0.502 & 0.634  & 1.054 \\
        1 & 3  & {16.27} & \textbf{0.425} & \textbf{0.495} & \textbf{0.611} & \textbf{0.991} \\
        \bottomrule
    \end{tabular}
    }

  \end{minipage}
  \vspace{-1em}
\end{table}

\subsection{Ablation for Memory Size}
Table~\ref{tab:ablation_distill_context} evaluates the effect of different memory sizes. Using a larger spatial memory size leads to slightly better PSNR metric, while a larger temporal memory size better preserves the pretrained model’s temporal continuity, resulting in better overall performance. Moreover, a larger spatial memory size may significantly increase the teacher model's memory size, as the spatial memory of adjacent chunks may completely differ, while their temporal memory overlaps. This not only increases the difficulty of training the teacher model but also poses challenges for distillation.

\begin{figure*}
  \centering
  \includegraphics[width=1\linewidth]{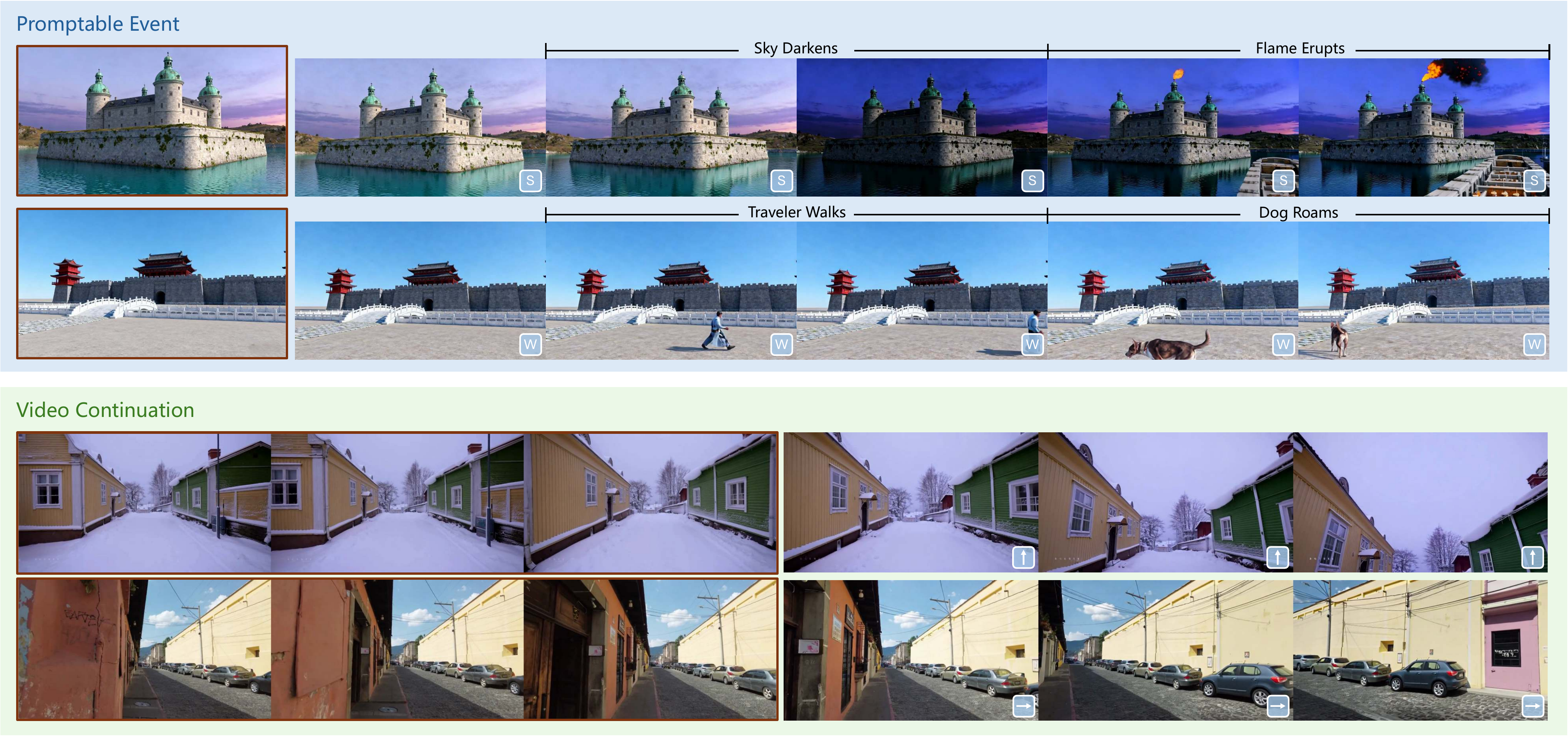}

  \caption{Visualization of promptable event and video continuation.
  }
  \label{fig:application}
  \vspace{-6mm}
\end{figure*}

\subsection{Evaluation on VBench}
We evaluate our model on VBench \cite{huang2024vbench} across diverse metrics. For each baseline, we provide the same image and action to generate long-horizon videos. The results presented in Fig. \ref{fig:vbench_plot} demonstrate the superior performance of WorldPlay. Notably, our method achieves outstanding results in key aspects such as consistency, motion smoothness, and scene generalizability.

\subsection{Evaluation on WorldScore}
We conduct a comprehensive evaluation using the WorldScore \cite{duan2025worldscore} benchmark, which consists of 2,000 diverse test cases encompassing various styles and scenarios for the static setting. Each case requires generating content based on an input image, a text prompt, and a specific camera trajectory. To assess the effectiveness of our model, we compare it against several state-of-the-art 3D and video generation baselines. Following the evaluation protocol established by Voyager \cite{huang2025voyager}, we focus on metrics that reflect controllability and generation quality across novel views. As demonstrated in Table \ref{tab:worldscore}, our method achieves the highest average score among all compared models, highlighting our model's superior performance in both precise controllability and high-fidelity generation quality.

\begin{table*}[t]
\footnotesize
\centering
\caption{Quantitative comparison on the \textit{WorldScore}. \textbf{\underline{Bold and underline}} presents the 1st, \textbf{Bold} indicates the 2nd, and \underline{underline} means the 3rd.}
\scalebox{0.9}{
\begin{tabular}{l|c|ccc|cccc}
\toprule
\textbf{Method} & \makecell{WorldScore \\ Average} & \makecell{Camera \\ Control} & \makecell{Object \\ Control} & \makecell{Content \\ Alignment} & \makecell{3D \\ Consistency} & \makecell{Photometric \\ Consistency} & \makecell{Style \\ Consistency} & \makecell{Subjective \\ Quality} \\
\Xhline{0.5pt}
WonderJourney \cite{yu2023wonderjourney} &  {63.75}&  {84.60}&  37.10 &  35.54&  80.60 & 79.03& 62.82 & \underline{66.56} \\
WonderWorld \cite{yu2025wonderworld} & \underline{72.69}&  \textbf{\underline{92.98}}&  51.76&  \textbf{\underline{71.25}}&  \textbf{\underline{86.87}}&  85.56& 70.57& 49.81 \\
EasyAnimate \cite{xu2024easyanimate}  &  52.85&  26.72&  54.50 &  50.76&  67.29&  47.35&  {73.05}& 50.31  \\
Allegro \cite{allegro2024} &  55.31&  24.84&  {57.47}&  {51.48}&  70.50 &  69.89&  65.60 &  47.41  \\
Gen-3 \cite{gen3} &  60.71&  29.47&  \underline{62.92}&  50.49&  68.31&  \underline{87.09}& 62.82 & {63.85}\\
CogVideoX-I2V \cite{yang2024cogvideox} & 62.15&  38.27&  40.07&  36.73&  \underline{86.21}&  \textbf{{88.12}}&  \underline{83.22}& 62.44\\
{Voyager} \cite{huang2025voyager} &  \textbf{{77.62}}&  \underline{85.95}&  \textbf{{66.92}} &  \textbf{68.92}&  {81.56}&  {85.99}& \textbf{{84.89}}& \textbf{{71.09}}\\
\hline
\textbf{Ours} &  \textbf{\underline{79.74}}&  \textbf{88.76}&  \textbf{\underline{69.05}} &  \underline{66.51}&  \textbf{86.43}&  \textbf{\underline{89.07}}& \textbf{\underline{85.17}}& \textbf{\underline{73.16}}\\

\bottomrule
\end{tabular}
}
\label{tab:worldscore}
\vspace{-0.5em}
\end{table*}

\section{User Study}

\begin{wrapfigure}{r}{0.5\textwidth}
  \centering
  \vspace{-10mm}
  \includegraphics[width=0.48\textwidth]{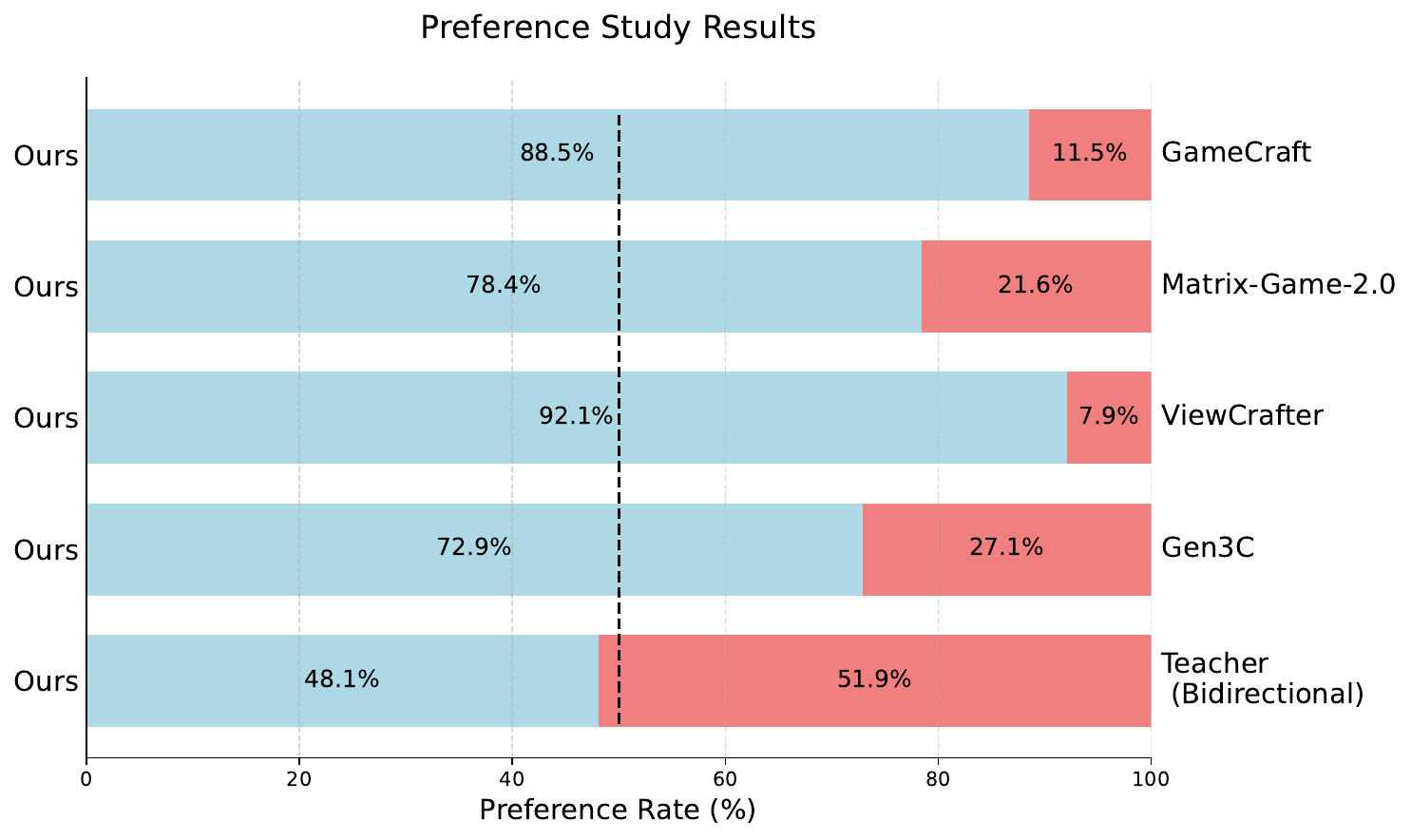}
  \caption{\textbf{Human evaluation.}}
  \label{fig:user_study}
  \vspace{-6mm}
\end{wrapfigure}

We conduct a comprehensive user study across multiple dimensions, including visual quality, control accuracy, and long-term consistency. In our setup, users are presented with two videos, generated from the same initial image and action inputs, and asked to select their preference based on the specified criteria. To ensure the robustness of our evaluation, we select 300 cases from diverse benchmarks such as VBench \cite{huang2024vbench} and WorldScore \cite{duan2025worldscore}, and 300 customized trajectories. The final results are then evaluated by a panel of 30 assessors. As shown in Fig. \ref{fig:user_study}, compared to other baselines, our distilled model achieves superior generation quality across all aforementioned evaluation metrics, clearly demonstrating our model's capability for both real-time interaction and long-term consistency.

\section{Additional Applications}

\subsection{Promptable Event}

Due to the autoregressive nature of WorldPlay, we can modify the text prompt at any time to control the subsequent generated content. Specifically, inspired by LongLive~\cite{yang2025longlive}, we employ a KV-recache technique to refresh the cached key–value states whenever the text prompt is modified. This effectively erases residual information from the previous prompt while preserving the motion and visual cues necessary to maintain temporal continuity. As shown in the upper part of Fig.~\ref{fig:application}, we can change the weather and trigger a fire eruption, or introduce new objects and characters. Through promptable event, we can generate various complex and uncommon scenarios, which can benefit agent learning by enabling agents to handle these unexpected situations.

\subsection{Video Continuation}
As shown at the bottom of Fig.~\ref{fig:application}, WorldPlay can generate follow-up content that remains highly consistent with a given initial video clip in terms of motion, appearance, and lighting. This enables stable video continuation, effectively extending the original video while preserving spatial-temporal consistency and content coherence, which opens up new possibilities in creative video generation and virtual environment construction.


\end{document}